\pdfoutput=1 
\documentclass[journal]{IEEEtran}

\usepackage{graphicx} 
\usepackage{epstopdf}

\usepackage{multirow}
\usepackage{algorithmic}
\usepackage{algorithm}
\usepackage{amsmath} %
\usepackage{amsthm}
\usepackage{amssymb}
\usepackage{amsxtra}
\usepackage{url} 
\usepackage{color} 
\usepackage{bm}
\usepackage{placeins}
\usepackage[caption=false]{subfig}
\usepackage{xspace}
\usepackage{rotating}
\usepackage[load-configurations = abbreviations]{siunitx}
\usepackage{textcomp}%

\newtheorem{theorem}{Theorem}[section]

\newtheorem{proposition}[theorem]{Proposition}

\newtheorem{mydef}{Definition}[section]

\newenvironment{definition}[1][Definition]{\begin{trivlist}
		\item[\hskip \labelsep {\bfseries #1}]}{\end{trivlist}}

\newcommand{\ie}{i.e.,\xspace}

\newcommand{\myvector}[1]{\bm{#1}}
\newcommand{\myvec}[1]{\myvector{#1}}
\newcommand{\pder}[2]{\frac{\partial#1}{\partial#2}}
\newcommand{\ppder}[3]{\frac{\partial^2#1}{\partial#2 \partial#3}}

\newcommand{\R}[1]{\mathbb{R}^{#1}}

\newcommand{\argmin}{\operatornamewithlimits{arg min}}
\newcommand{\argmax}{\operatornamewithlimits{arg max}}

\newcommand{\velocity}[1]{\SI[per-mode=symbol]{#1}{\meter \per \second}}
\newcommand{\accel}[1]{\SI[per-mode=symbol]{#1}{\meter \per \second \squared}}
\newcommand{\dur}[1]{\SI{#1}{\second}}
\newcommand{\dist}[1]{\SI{#1}{\meter}}
\newcommand{\distcm}[1]{\SI{#1}{\centi\meter}}

\newcommand{\huv}[1]{\myvec{\hat{a}_{#1}}}
\newcommand{\huvp}{\myvec{\hat{a}}}
\newcommand{\uvv}[1]{\myvec{a_{#1}}}
\newcommand{\hu}[1]{\hat{a}_{#1}}

\newcommand{\Lvp}{\myvec{\Gamma}}

\newcommand{\usample}[1]{a_{i{#1}}}

\newcommand{\C}{\myvec{C}}

\newcommand{\qifunctionZ}[1]{\qifunctionpointZ{#1}{a}}
\newcommand{\qifunctionpointZ}[2]{Q_{\x , #1 }^{( \myvec{0} )}(#2)}

\newcommand{\qcoefs}[1]{\myvec{p}_{#1}}
\newcommand{\qcoefsi}{\qcoefs{i}}

\newcommand{\qsampleZ}[1]{Q_{\x , #1}(\usample{#1})}

\newcommand{\noise}[1]{\myvec{\eta}_{#1}}

\newcommand{\noiseMu}[1]{\mu_{\noise{#1}}}

\newcommand{\noiseStd}[1]{\sigma_{\noise{#1}}}

\newcommand{\gauss}[2]{\mathcal{N}(#1 , #2)}
\newcommand{\gaussNoise}[1]{\gauss{\noiseMu{#1}}{\noiseStd{#1}}}

\newcommand{\policyApp}{\myvec{\hat{\pi}}(\x)}

\newcommand{\bnx}[1]{\myvec{f_1}(\x_{#1},\noise{#1})}
\newcommand{\oo}[1]{\mathcal{O}(#1)}

\newcommand{\frm}{PEARL}
\newcommand{\fullName}{PrEference Appraisal Reinforcement Learning}

\newcommand{\pbt}{PBT}

\newcommand{\mdp}{MDP}
\newcommand{\rl}{RL}

\newcommand{\X}{S}
\newcommand{\x}{\ensuremath{\myvec{s}}}
\newcommand{\dx}{\ensuremath{\dot \x}} 
\newcommand{\ddx}{\ensuremath{\ddot \x}} 

\newcommand{\U}{A}
\newcommand{\uv}{\ensuremath{\myvec{a}}}
\newcommand{\ui}[1]{a}

\newcommand{\teta}{\ensuremath{\myvec{\theta}}}
\newcommand{\F}{\ensuremath{\myvec{F}}}
\newcommand{\Fi}{F}

\newcommand{\D}{\ensuremath{\mathbf{D}}}
\newcommand{\f}{\ensuremath{\myvec{f}}}
\newcommand{\g}{\ensuremath{\myvec{g}}}
\newcommand{\A}{\ensuremath{\myvec{g}}(\x)}
\newcommand{\bx}{\ensuremath{\myvec{f}}(\x)}

\newcommand{\nr}{\ensuremath{{d_r}}}
\newcommand{\na}{\ensuremath{d_a}}
\newcommand{\nx}{\ensuremath{d_s}}

\newcommand{\no}{\ensuremath{n_p}}

\newcommand{\ev}[1]{\myvec{e_{#1}}}
\newcommand{\obj}[1]{{\myvec{p}}_{#1}}
\newcommand{\proj}[2]{{\myvec{\text{pr}}}_{#2}^{\obj{#1}}(\x)}

\newcommand{\ac}{\uv}

\usepackage[normalem]{ulem}
\usepackage{tabularx}

\begin{document}
	
	\title{PEARL: PrEference Appraisal Reinforcement Learning for Motion Planning}

	\author{Aleksandra Faust,
		Hao-Tien Lewis Chiang,
		and~Lydia Tapia
		\thanks{A. Faust is with Google Brain, Mountain View, CA 94043, USA e-mail: faust@google.com.}%
		\thanks{H. Chiang, and L. Tapia are with Department of Computer Science, University of New Mexico, Albuquerque, NM, USA.}
	}			
	\maketitle
	\begin{abstract}
		Robot motion planning often requires finding trajectories that balance different user intents, or preferences. 
        One of these preferences is usually arrival at the goal, while another might be obstacle avoidance. Here, we formalize these, and similar, tasks as preference balancing tasks (PBTs) on acceleration controlled robots, and propose a motion planning solution, PrEference Appraisal Reinforcement Learning (PEARL). PEARL uses reinforcement learning on a restricted training domain, combined with features engineered from user-given intents. PEARL's planner then generates trajectories in expanded domains for more complex problems. We present an adaptation for rejection of stochastic disturbances and offer in-depth analysis, including task completion conditions and behavior analysis when the conditions do not hold. PEARL is evaluated on five problems, two multi-agent obstacle avoidance tasks and three that stochastically disturb the system at run-time: 1) a multi-agent pursuit problem with 1000 pursuers, 2) robot navigation through 900 moving obstacles, which is is trained with in an environment with only 4 static obstacles, 3) aerial cargo delivery, 4) two robot rendezvous, and 5) flying inverted pendulum. Lastly, we evaluate the method on a physical quadrotor UAV robot with a suspended load influenced by a stochastic disturbance. The video, \url{https://youtu.be/ZkFt1uY6vlw} contains the experiments and visualization of the simulations.
	\end{abstract}
	
	
	\maketitle
	\section{Introduction}
	\label{sec:intro}
	
	\begin{figure}[t]
		\centering
		\subfloat[Multi-agent pursuit]{\includegraphics[width=0.5\textwidth,height=40mm,keepaspectratio=true]{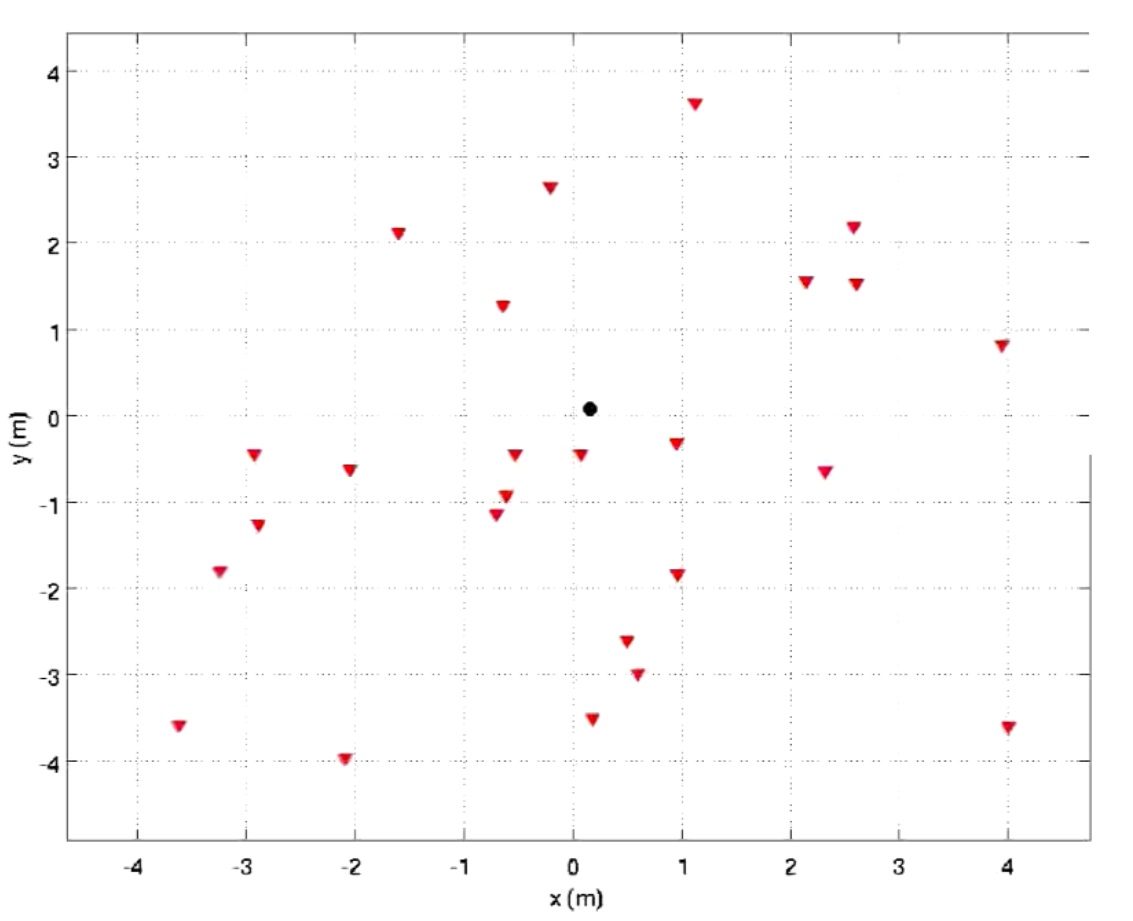}\label{fig:multiagent}}
		
		\subfloat[Dynamic obstacle avoidance]{
			\includegraphics[trim=13mm 13mm 13mm 13mm,clip,width=0.33\textwidth,height=40mm,keepaspectratio=true]{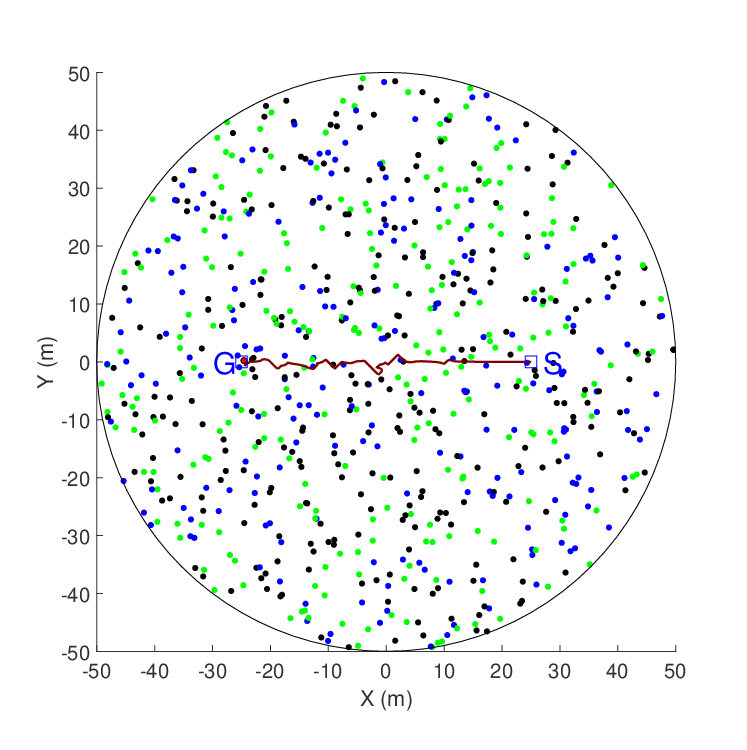}
			\label{fig:obstacle}\label{fig:pathObsRobotAndEnv}}
		\subfloat[Aerial cargo delivery]{\includegraphics[width=0.33\textwidth,height=40mm,keepaspectratio=true]{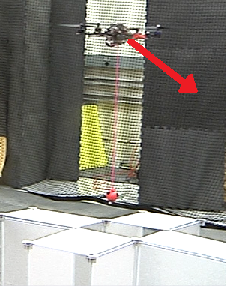}\label{fig:swingfree}}		
		
		\subfloat[Flying inverted pendulum]{\includegraphics[width=0.33\textwidth,height=30mm,keepaspectratio=true]{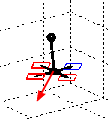}
			\label{fig:invPend}}
		\subfloat[Rendezvous]{\includegraphics[width=0.33\textwidth,height=30mm,keepaspectratio=true]{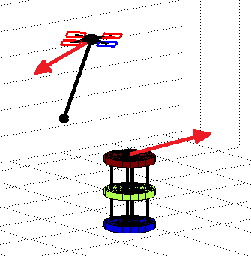}\label{fig:rand}}
		\\
		\caption{Preference-balancing task examples: a) a multi-agent pursuit system where pursuers (red triangles) chase a prey (black circle) while avoiding collision, b) an agent navigating (red line) to a goal (G) while avoiding up to 900 moving obstacles with varied dynamics (colored squares), c) a quadrotor delivering a suspended load with minimal load disturbance, d) 3D navigation while balancing a pole, and e) joint planning of a quadrotor and mobile robot to transfer a suspended load with minimal disturbances. Red arrows in c), d) and e) represent external input stochastic disturbance acting on the robot.}
		\label{fig:examples}
	\end{figure}

	Motion planning in complex scenarios, like multi-robot coordination (Figures \ref{fig:multiagent} and \ref{fig:rand}), dynamic obstacle avoidance (Figure \ref{fig:obstacle}), aerial cargo delivery (Figure \ref{fig:swingfree}), or balancing a flying inverted pendulum (Figure \ref{fig:invPend}), require finding trajectories for systems with unknown non-linear dynamics. To complete these tasks, it is not sufficient that the planned trajectories are optimized only for speed, but they need to balance different, often opposing, qualities (preferences) along the way.  In some cases the preferences can be formulated as constraints on the trajectory, but often 
	the precise constraints are unknown or difficult to calculate. For example, consider a simple manipulation task where a robot is required to set a glass on a table without breaking it. We do not know precisely the amount of force that causes the glass to shatter, yet we can describe our preferences: low force and fast task completion.  To find feasible planning solutions, these complex problems can frequently be described in terms of points of interest, or intents, that the agent either needs to progress towards (attractors) or stay away from (repellers). The planning must balance between them. We call these tasks Preference Balancing Tasks (PBTs). There are several challenges associated with solving \pbt s: finding a policy that guides the agent to the goal, handling non-linear dynamics, and adaption to external disturbances in real-time. 
	
	An example of a  solution to a simple PBT is an
	Artificial Potential Field (APF) \cite{lccf-hcrntahhrc-11}, which balances a force that attracts the robot to the goal with a force that repulses the robot from obstacles. These two forces can be thought of as preferences that drive the robot motion through the environment. The challenge of planning with methods such as APFs, is primarily in determining the placement, shape, and relative scaling between the potentials. The properly constructed force field greatly impacts the method's success. 
	
	Reinforcement learning (RL) has been recently successful in planning trajectories for systems in unknown dynamics \cite{kober_IJRR_2013}. RL often learns a state-value function, which is used much like APF to plan a trajectory. In robotics domains, RL requires function approximation, either with Deep Neural Nets or carefully selected features \cite{BusBab:10-002}. Deep reinforcement learning (DRL), RL with deep neural networks as function approximators, gained lots of attention recently in applications such as Atari games \cite{atari-paper}, self-driving \cite{chen2015deepdriving}, and learning motion planning policies from raw sensor data \cite{tamar2016value} \cite{prm-rl}. 
	Deep neural nets are good choice of approximators when we do not have good intuition about the problem feature space, and have abundant data for training.
	At the same time, the DRL methods have been rather challenging to apply to robotics motion planing, due to the speed of decision-making, the training data needed, and the general instability of training \cite{sim2real}.
	
	In contrast, \frm, as well as other feature-based RL methods, is more interpretable, requires less data, and executes faster. 
	\frm\ is appropriate for problems where we have an intuition of the feature set, lack the demonstration data to use for training, and require fast training. While DRL takes hours and days to train, \frm\ completes training within minutes. Additionally, \frm's features make the behavior easy to interpret and analyze. Furthermore, for tasks with only attractors, we know the conditions under which the task's goal is an asymptotically stable point \cite{faust-acta-13}, which means that we can guarantee the agent's behavior when the conditions are met. In Section \ref{sec:discussion}, we show that the proposed features satisfy the stability criterion for attractor-only multi-agent tasks. And for tasks with mixed intents (both attractors and repellers), we analyze the local minima conditions.  
	
	External disturbances that influence a robot's motion at runtime (e.g., atmospheric changes or wind) pose another challenge. They are often stochastic and can externally excite the system with normally distributed intensity and direction, with variation between consecutive observations \cite{asstrom-stochcontrol}. 
	Stochastic disturbances, along with complex nonlinear system dynamics, make traditional solutions, e.g., adaptive and robust control modeling, which solve this problem by explicitly solving the optimal control problem, difficult or intractable \cite{Khalil-nonlinear-96}. We are interested in a trajectory generation method that solves \pbt, rejects stochastic disturbances, and is computationally efficient so as to be used on a high-dimensional system that requires frequent input selection (i.e., 50 Hz). 
	
	We propose Preference Appraisal Reinforcement Learning (PEARL), a simple and easy method to solve many PBTs for multi-agent systems with unknown non-linear dynamics. PEARL constructs a RL state-value function (APF equivalent) from user-given intents as points of interest (preferences), computes features from the intents, and learns the scaling between the features (appraisal). We learn the state-value function on simplified problems to speed up the training, and use it to plan more difficult problems. To make the trajectory planning robust to external disturbances, we present a policy approximation, Least Squares Axial Policy Approximation (LSAPA) \cite{faust-icra-15}, that rejects stochastic disturbances. PEARL trains in minutes, and although there are no guarantees that it will work in all cases, we show that it works on a range of problems and offer an analysis of the method's limitations.  
	
	We evaluate PEARL with and without stochastic disturbances, on dynamic obstacle avoidance \cite{faust-icra-16}, aerial cargo delivery (Figure \ref{fig:swingfree}) \cite{faust-icra-15}, rendezvous (Figure \ref{fig:rand}) \cite{faust-icra-15}, and flying inverted pendulum (Figure \ref{fig:invPend}) tasks. Further, PEARL was used for a multi-agent pursuit problem to plan in 100-dimensional state space in real-time at 50 Hz.  The feasibility of the generated trajectory is verified experimentally for the aerial cargo delivery task with stochastic disturbances \cite{faust-icra-15}. 
	
	This paper extends our prior presentation of a single-agent \frm\ \cite{faust-icra-16} and stochastic disturbance rejection \cite{faust-icra-15} to the multi-agent platform, along with comprehensive analysis and evaluation that shows \frm\ flexibility and range. 
	New to this paper are 1) multi-agent formulation of PEARL, 2) training domain formalization and analysis, 3) stability analysis, 4) multi-agent pursuit solution, and 5) flying inverted pendulum under stochastic disturbances. 
	
	The rest of the paper is structured as follows. Section \ref{sec:relwork} positions \frm\ within the related work. Section \ref{sec:Preliminiaries} introduces necessary preliminaries. In Section \ref{sec:methods} we present \frm's components: preference balancing tasks, the MDP and training domain, construction of features, fittest policy selection, and policy adaptation for rejection of stochastic disturbances. Section \ref{sec:results} presents comprehensive case studies demonstrating the method. In Section \ref{sec:discussion} we analyze the method, its progression to the goal, computational cost, feature properties, and construction of the learning domain. Lastly, we conclude in Section \ref{sec:conclution}.   
	
	\begin{figure}[tb]
		\begin{center}
			\includegraphics[width=0.5\textwidth]{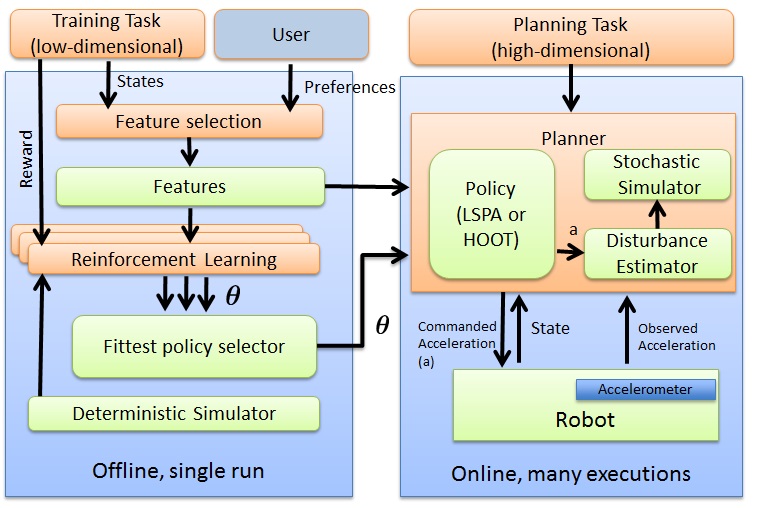}
		\end{center}
		\caption{\small \fullName\ (\frm ) framework for learning and executing \pbt .
			The user intents (preference), given as a point and type, are encoded into features. 
			The learning agent appraises feature priorities with reinforcement learning on a low-dimensional training
			problem. The outputs of the learning are feature weights, $\theta$. To learn the weights offline, we use a deterministic simulator, do learning in several rollouts, and keep the best weights. After the weights are computed, we deploy a corresponding policy online.
			At every planning iteration step, we select the acceleration to apply to the robot based on the current robot state, learned weights, and selected features. Different action selection policies can we used. For stochastic disturbances, we equip a robot with an accelerometer. The disturbance estimator computes disturbance and variance based on the observed and commanded acceleration. The stochastic simulator generates a stochastic disturbance with the given parameters when estimating the next steps.
		}
		\label{fig:warf14Arch}       
		\label{fig:flow}
	\end{figure}
	\section{Related work}
	\label{sec:relwork}
	
	\emph{Reinforcement learning: } Function approximation RL methods typically assume user-provided features
	\cite{BusBab:10-002}, which map subspaces to points. RL is
	sensitive to feature selection, because of the implied dimensionality reduction \cite{BusBab:10-002}. Classically, two
	feature types are used in RL: discretization and basis functions
	\cite{wu-largescale-10}. Discretization partitions the domain, scaling
	exponentially with the space dimensionality. Basis functions, such as kernels
	and radial basis function networks, offer more learning flexibility. These functions, however, can
	require manual parameter tuning, and the feature number  for multi-robot systems and dynamic obstacle avoidance tasks  increases exponentially
	with the state space dimensionality \cite{wu-largescale-10}. Further, these method are generic and do not capture prior knowledge of a task that an engineer might have. \frm\ proposes a feature
	selection method that solves a particular class of motion tasks for
	acceleration-controlled robots, exploiting task knowledge. The
	number of features is invariant to the problem dimensionality, and the computation time
	scales polynomially with the state space dimension. Similar to Voronoi
	decomposition, which solves high-dimensional manipulation problems by projecting the
	robot's configuration space onto a low dimensional task space
	\cite{shkolnik-highDim-09},
	the features we propose define a basis in the preference-task space as well.
	In contrast to Voronoi decomposition, \frm\  learns the relationship between the features.
	Another RL approach solves manipulation tasks with hard \cite{stilman-hard-2010}
	and soft \cite{kunz-soft-2012} constraints. Our tasks, however, do not have
	known constraints and bounds; they are set up as preferences to guide
	dynamically feasible trajectory generation. 
	
	\emph{Preference shaping:}
	Devlin et al. \cite{devlin-rewards-aamas14} focuses on reward shaping to improve the equity of the joint agent
	policy in multi-agent tasks. Rather, we put emphasis on
	the feature selection and task formulation. Another approach dynamically
	calculates preference ordering where preferences are expressed as logical
	formulas \cite{padgham-preferences-aamas2013}. Our preferences adapt easily to spaces of different dimensions.
	
	\emph{Stochastic disturbances:} Robot motion control under stochastic disturbances has been studied on a number of problems, including quadrotor trajectory tracking under
	wind-gust disturbances 
	\cite{alexis-constrained-10}, 
	a blimp path planning  in the presence of stochastic wind 
	\cite{kawano-blimp0-11}, with different approaches 
	use low-level controllers for stabilization of
	trajectories within reach tubes \cite{decastro-guaranteeing-13} or trajectory libraries \cite{majumdar-robust-13}. Other approaches to UAV control in environments with a drift field explicitly solve the system dynamics \cite{Masoud-potential-11, shen-stochQuad-12}, or use iterative learning to estimate repetitive disturbance \cite{schoellig-iterative-12, muller-mpc-13}.  While these solutions are aimed at particular systems or repetitive disturbances, our method works for a class of problems and systems influenced by the stochastic disturbance. Finally, Nonlinear Model Predictive Control (NMPC) \cite{nmpc-2011} tracks a deterministic trajectory minimizing a cumulative error for a non-linear dynamical system in presence of the stochastic disturbances. The main difference is that, while NMPC computes the cost on-the-fly, PEARL constructs a state-value function, a discounted, infinite-horizon, cumulative cost (reward) prior to task execution. PEARL's control policy, LSAPA, uses the state-value function during planning and only needs to solve a one-step optimization problem with respect to the state-value function.
	
	\emph{Continuous action policy approximation:}
	Several sampling-based methods that approximate the policy efficiently exist, 
	such as HOOT
	\cite{mansley-hoot-11}, which uses hierarchical discretization to progressively narrow the search on the
	most promising areas of the input space, thus ensuring an arbitrarily small error
	\cite{mansley-hoot-11}. 
	Deterministic Axial Sum (DAS) \cite{faust-acta-13} solves input selection in linear time using divide and conquer. It finds optimal input for each of the input axes independently with Lagrangian interpolation, and then combines the single-axis selections. 
	DAS works well for high-dimensional multi-agent planning with value functions without local maxima. HOOT is slower, and, in practice, HOOT works well for single-agent planning with value functions that have many small-scale maxima. Although DAS can compensate for some levels of zero-mean noise \cite{faust-acta-13, figueroa-wcica-14}, the method stops working in the presence of stochastic disturbances. This is because the external disturbance induces unpredictable drift onto the system. The method presented here, LSAPA, uses polynomial least squares regression instead of interpolation to compensate for the stochastic disturbances.
	
	\section{Background}
	\label{sec:Preliminiaries}
	
	We model
	robots as discrete-time nonlinear control-affine systems with accelerations as control inputs
	\cite{LavalleBook06}.
	Consider a robot with $m$ degrees of freedom (DoF). If an acceleration $\ac(n)
	\in \R{m}$ is applied to the robot's center of mass at time-step $n$, the new
	position-velocity vector (state) $\x(n+1) \in \R{2m}$ is,
	\begin{equation}
	D: \;\;\;\;\;\x(n+1) = \myvec{f}(\x(n)) + \myvec{g}(\x(n))\ac(n),
	\label{eq:dynSystem}
	\end{equation}
	for some functions $\myvec{f}, \myvec{g}$. 
	A \textit{Markov decision process} (MDP), a tuple $(\X, A, D, R)$ with states
	$\X \subset \R{2m}$ and actions $A \subset \R{m}$, that assigns immediate scalar rewards
	$R:\X\rightarrow\R{}$ to states in $\X$, formulates a task for the system
	\eqref{eq:dynSystem} \cite{BusBab:10-002}. A solution to a MDP is a control
	policy $\pi :\X \rightarrow A$ that maximizes cumulative discounted reward over the 
	agent's lifetime, value, $V(\x(0)) = \sum_{i=0}^\infty \gamma^i R(\x(i))$, where
	$0 \leq \gamma \leq 1$ is the discount factor. 
	\textit{Approximate value iteration} (AVI) \cite{ernst-avi}, finds a solution to
	a continuous state MDP by approximating state-value function $V$ with a linear
	map
	\begin{equation}
	\label{eq:value}
	\hat V(\x) = \myvec{\theta}^T\myvec{F}(\x). 
	\end{equation}
	RL often works with \textit{action-value function}, $Q:\X \times \U \rightarrow \R{}$, a measure of the discounted accumulated reward collected when an action $\uv$ is taken at state $\x$ \cite{bartoBook98}. In relation to the state-value function, V \eqref{eq:value}, action-value $Q$ can be expressed as
	\begin{equation}
	\label{eq:q}
	Q(\x, \uv) =  V(D(\x,\uv)) = \sum_{i=1}^{\no} \theta_i \myvec{F_i}(D(\x,\uv)).
	\end{equation}
	AVI takes a feature vector $\myvec F(\x)$ and learns weights $\myvec \theta$
	between them by sampling the state-space and observing the rewards. It
	iteratively updates $\teta$ in an expectation-maximization manner.
	
	After parameter learning is completed, batch RL enters a planning phase. The
	planner takes the value function approximation \eqref{eq:value} and an initial
	condition, and it generates a trajectory using the closed-loop control with a greedy policy
	with respect to the state-value approximation,
	\begin{equation}
	\label{eq:greedy}
	\pi^{\hat V}(\x) = \argmax_{\ac \in A} \hat V(\x')
	\end{equation}
	where state $\x'$ is the result of applying action $\ac$ to state $\x$. 
	Action selection in continuous spaces, which calculates the greedy policy
	\eqref{eq:greedy}, is a multivariate optimization over an unknown function.
	
	\section{PEARL}
	\label{sec:methods}
	\frm\ solves a \pbt\ in two phases, offline learning in simulation, 
	and planning, described in Sections \ref{sec:training} and \ref{sec:planning}.	
	Figure \ref{fig:warf14Arch} shows the \frm\ components and flow.
	
	To start the learning phase, a user provides \frm\ with the task definition, which contains the  basic information about the problem: the number of robots,
	the robot's DoFs, maximum accelerations, and a set of intents (attractors or repellers). The basic system information is encoded into a training MDP (Section \ref{sec:mdpsetup}). The intents are encoded into features (Section \ref{sec:feature}). To make the learning more tractable, we select a learning domain that is smaller than the full problem space, (Section \ref{sec:learning_mdp}). The features and training MDP are passed to the approximate value iteration algorithm (AVI \cite{ernst-avi} or CAFVI \cite{faust-acta-13}), which determines the weights between features. The training output is a feature weight vector $\myvec{\theta} \in \R{\no},$ where $\no$ is number of preferences. We repeat the training several times to select the best policy (Section \ref{sec:learning_mc}). 
	
	Once the fittest policy is selected, \frm\ is ready to execute tasks on the full MDP domain. The features, weights, and initial conditions are passed to the planner (Section \ref{sec:planning}), which uses a greedy policy with respect to the learned state-value function to complete the task. Section \ref{sec:lspa} presents a greedy policy modification that allows the system to compensate for stochastic disturbances in the special case when all task preferences are attractors.  
	
	\subsection{Training}
	\label{sec:training}
	Our aim is to solve tasks that can be described with a set of
	attractors (goals) and repellers (obstacles) for acceleration-controlled multi-robot systems with unknown dynamics.  
	
	\subsubsection{Multi-agent MDP Setup}
	\label{sec:mdpsetup}
	The multi-robot system consists of $\nr$ robots, where $i^{th}$ robot has  ${\nr}_i$ degrees of freedom (DoF). We assume the robots work in continuous state and action spaces, are controlled through acceleration applied to their center of mass, and have differentially flat dynamics that are not explicitly known. Let ${\x}_i,\; \dot {\x}_i\; \in \R{{\nr}_i}\; \ddot {\x}_i \in \R{{\na}_i}$ be the $i^{th}$ robot's position, velocity, and acceleration, respectively. The \mdp\ state space is $\X \subset 2\R{\nx},$ where $$\nx = \sum_{i=1}^{\nr} {\nr}_i$$ is the dimensionality of the joint multi-robot system. The state $\x \in \X$ is joint vector 
	$$\x = [{\x}_1,...,{\x}_{\nr},{\dx}_1,..., {\dx}_{\nr}]^T,$$ 
	and action $\ac \in A=\R{m}$ is the joint acceleration vector, $$\ac = [{\ddx}_1,..., {\ddx}_{\na}]^T.$$ 
	The state transition function, which we assume is unknown, is a joint dynamical system of individual robots $$\D = {\D}_1 \times ... \times {\D_{\nr}},$$ 
	each being control-affine \eqref{eq:dynSystem}. Note that since all robots are control-affine, the joint system is as well. 
	
	For training purposes, we assume the presence of a black-box simulator, or dynamics samples, for each robot. The reward $R$ is set to one when the joint multi-robot system achieves the goal, and zero otherwise. The tuple
	\begin{equation}
	\label{eq:mdp}
	\mathcal{M} = (\X, \U, \D, R)
	\end{equation}
	defines the joint MDP for the multi-robot problem.
	
	\subsubsection{Feature Selection}
	\label{sec:feature}
	We define a \pbt\ with $\no$ preferences,
	$P = \obj{1},...,\obj{\no}$. The preferences, points in position or velocity space, $\obj{i} \in \R{\nr_i},\,i=1,..,\no$, either attract or repel the one or more agents. Preferences that attract an agent are goals, $P_a$, whereas the preferences that repel it are obstacles, $P_r,$ $P = P_a \cup P_r,\, P_a \cap P_r = \emptyset$. We assume that the task is well-defined, i.e., the task's attractors forms a non-empty region, not fully occluded by repellers, $\{\cap_{\obj{} \in P_a} \obj{} \} \setminus \{\cap_{\obj{} \in P_r} \obj{} \} \ne \emptyset$. 
	
	To learn a \pbt~with $\no$ preferences, $\obj{1},...,\obj{\no}$, we form a feature
	for each preference. 
	We construct the features, associated with both attractors and repellers, and reduce their measure to the intended point of interest. 
	For example, multi-agent pursuit has two attractors, the prey's position and speed, and a repeller, other nearby agents' positions. The three preferences form three features that intuitively correspond to the distance from the prey, speed difference from the prey, and the inverse of the distance from other agents. 
	
	Formally, assuming the low-dimensional task space and high-dimensional
	\mdp\ space $\no \ll \nx$, we consider \textit{task-preference features}, 
	\begin{equation}
	\label{eq:f}
	\F(\x,\nx) =  [F_1(\x,\nx),...,F_{\no}(\x,\nx)]^T.
	\end{equation}
	Parametrized with the state space dimensionality, $\nx$, the features map the
	state space $\X$ to a lower dimensional feature space, and, depending on the preference type,
	measure either the squared intensity or distance to the attractor. 
	Let 
	$S_i 	\subset \{1,..,\nr\}$ be a subset of robots that a preference $\obj{i}$ applies to, and 
	$\proj{i}{j}$ be a projection of the $j^{th}$ robot's state onto the minimal
	subspace that contains $\obj{i}$. For instance, when a preference $\obj{i}$ is a
	point in a position space, $\proj{i}{j}$ is the robot's position.
	Similarly, when $\obj{i}$ is a point in a velocity space, $\proj{i}{j}$ is the robot's velocity. Then, \textit{attractor features} are defined
	with
	\begin{equation}
	\label{eq:distance}
	\Fi_i(\x,\nx) = \sum_{j \in S_i} \|\proj{i}{j} - \obj{i} \|^2,
	\end{equation}
	and \textit{repeller features} are defined with \begin{equation}
	\label{eq:intensity}
	\Fi_i(\x,\nx) = \sum_{j \in S_i} (1+ \|\proj{i}{j} - \obj{i} \|^2)^{-1}.
	\end{equation}
	
	\subsubsection{Learning Efficiency}
	\label{sec:learning_mdp}
	
	\frm\ uses an AVI-based \rl\ algorithm  \cite{ernst-avi}, \cite{faust-acta-13} to discover the relative
	weights between the features (preference appraisal). 
	AVI algorithms learn by sampling from the learning domain. In high-dimensional and high-volume learning domains, learning algorithms need prohibitively many iterations, or might not even converge due to the curse of dimensionality. To avoid these problems and make learning computationally efficient, we propose a training domain \eqref{eq:mdp_l} that is a subset of the full problem domain $\mathcal{M}$  in \eqref{eq:mdp}. Let the training MDP be 
	\begin{equation}
	\label{eq:mdp_l}
	\mathcal{M}_l = (\X_l, \U_l, \D, R),
	\end{equation}
	where $\X_l \subseteq \X,$ and $\U_l \subseteq \U.$ The training MDP, $\mathcal{M}_l's$ domain, is a subset of the full training domain. Specifically, out of all possible subsets, we select $\X_l$ and $\U_l$ such that they contain all of the task's point of interest and its immediate vicinity, but not much more. For instance, in the case of dynamic obstacle avoidance (Figure \ref{fig:obstacle}), the problem domain is a circular area with radius 50 meters in 2-dimensional workspace, but if we place three obstacles at 2 meters from the goal, then the training domain can be a circle with radius 3 meters from the goal, since it contains the attractor and all the repellers, and all are reachable. To formalize, the training domain $\X_l$ is a bounded, closed set that contains $\mathcal{B},$ the smallest open set that covers all the objectives, 
	$$\mathcal{B} = \cap \{x\, |\, \exists \epsilon_i > 0, \|x- \obj{i} \| < \epsilon_i, i \in \{1,..,\no\}\}.$$
	The training set, $\X_l,$ is $\mathcal{B} \subset \X_l \subseteq \X$. This set is used in the RL training algorithm to sample the training tuples.
	
	\subsubsection{Monte Carlo Policy Selection}
	\label{sec:learning_mc}
	We repeat the training $n_{mc}$ times because of the sampling nature of RL algorithms. Each training trial produces a different set of preference weights, resulting in $\myvec{\theta}_i,$ for  $i \in 1,...,n_{mc}.$ To select the fittest policy, we evaluate each on a fixed evaluation set of initial conditions. The evaluation measures the percent success rate for the policy and the average trajectory duration. We select a policy with the fastest average trajectory from the policies with the highest success rates on the training domain.
	
	\subsection{Trajectory Planning}
	\label{sec:planning}
	After a \frm\ agent is trained, we use the learned value function to plan. 
	To plan and generate trajectories, the planner executes a closed-loop feedback control. At each time step $n,$ the agent observes the state of the world, $\x(n)$, and uses an approximation of a greedy policy \eqref{eq:greedy} such as Deterministic Policy Approximation (DAS) \cite{faust-acta-13} or HOOT \cite{mansley-hoot-11} to compute an action, $\uv,$ to apply to the system.
	
	\subsubsection{Stochastic Disturbance Rejection for Attractor Tasks}
	\label{sec:lspa}
	We look at the special case of the attractor-only tasks, e.g., rendezvous (Figure \ref{fig:rand}), flying inverted pendulum (Figure \ref{fig:invPend}), and aerial cargo delivery (Figure \ref{fig:swingfree}).  For these problems, we present a policy approximation that rejects stochastic disturbances that influence the system at run-time.
	The policy approximation, Least Squares Axial Policy Approximation (LSAPA), adapts DAS \cite{faust-acta-13}. In \cite{faust-acta-13}, we showed that for tasks with only attractors a Q-value function is a quadratic function of the state $\x$, therefore DAS can be used for efficient, high-precision planning if certain conditions are met. Specifically, DAS takes advantage of the facts that action-value function $Q$ is a quadratic function of the input $\uv$ for any fixed arbitrary state $\x$ in a control-affine system \eqref{eq:dynSystem} with state-value approximation \eqref{eq:value} \cite{faust-acta-13}. DAS finds an approximation for the maximum local $Q$ function for a fixed state $x$. It works in two steps, first finding maxima on each axis independently and then combining them together. To find a maximum on an axis, the method uses Lagrangian interpolation to find the coefficients of the quadratic polynomial representing the $Q$ function. 
	Then, an action that maximizes the $Q$ function on each axis is
	found by zeroing the derivative. The final policy is a piecewise maximum of a convex and simple vector sums of the action maxima found on the axes. 
	Deterministic axial policies do
	not adapt to changing conditions or external
	forces, because their results do not depend on the selected samples.
	We extend DAS to work in the presence of
	disturbances via LSAPA. LSAPA uses least squares regression, rather than Lagrangian
	interpolation, to select the maximum on a single axis. This change allows
	the LSAPA method to compensate for the error induced by non-zero mean disturbances.
	
	We consider a system that is externally influenced by disturbance
	\begin{equation}
	\label{eq:stoch_dyn}
	\D_s: \;\;\; \x_{k+1} = \f({\x_k}) + \g(\x_k)(\uv_k + \noise{k}),
	\end{equation}
	where $\noise{k}$ is sampled from a Gaussian distribution, $\gaussNoise{k}$, and show that a) the Q-function remains quadratic, and b) propose a modification to DAS that compensates for the external disturbance.
	
	During the planning, we assume that we have a black-box simulator
	of the system, which receives mean and variance of the current probability distribution of the
	disturbance $\gauss{\mu_k}{\sigma_k}$. The probability distribution can be obtained by estimating a moving average and variance of the error between observed and desired acceleration, obtained with an accelerometer placed on a robot. 
	
	At every time step, $k$, LSAPA, observes a state, $\x_k$. By sampling the simulator, LSAPA finds a near-optimal input, $\myvec{u}_k$, to apply to the system.
	The Lagrangian interpolation, used in DAS, 
	interpolates the underlying quadratic function with only three points, and this compounds the error from the disturbances. In contrast, our new method, LSAPA, uses least squares regression with many sample points to compensate for the induced error.

	\begin{table}
		\centering
		\caption{\small Summary of key symbols and notation.}
		\label{tab:symbols}
		\begin{tabular}{l|l}
			Symbol & Description\\
			\hline
			$Q:\X \times \U \rightarrow \R{}$    & Action-value function\\ 
			$\noise{k} ~\gaussNoise{k}$  & External force excreted onto the \\
			$\ev{n}$ & n$^\text{th}$ axis unit vector\\
			$\qcoefsi = [p_{2,i} \; p_{1,i}\; p_{0,i}]^T \in \R{3}.$ & Coefficients of Q's axial restriction\\
			$\uv \in U$    & Input vector\\ 
			$\ui{i} \in \R{}$    & Univariate input variable\\ 
			$\uvv{n} \in \R{}$    & Set of vectors in direction of n$^\text{th}$ axis\\             
			$\hu{n} \in \R{}$    & Estimate in direction of the n$^\text{th}$ axis\\             
			$\huv{n} = \sum_{i=1}^{n} \hu{n} \ev{i}$    & Estimate over first n axes\\             
			$\huvp$    & $Q$'s maximum estimate with a policy\\             
			$\qifunctionZ{n} = Q(\x, \myvec{p} + u\myvec{e_n})$ & Univariate function in the direction \\ & of axis $\myvec{e_n}$, passing through point $\myvec{p}$\\ 
			$d_n $ & Number of axis samples \\
			$\na $ & Input dimensionality \\
			
			\hline
		\end{tabular}
	\end{table}
	We first show that the $\myvec{Q}$ function remains quadratic with a maximum even when the system is influenced with a stochastic term.
	\begin{proposition}
		\label{th:prop}
		\label{th:concave}
		Action-value function $Q(\x, \uv)$ \eqref{eq:q} corresponding to state-value function $V$ \eqref{eq:value} and a discrete-time system \eqref{eq:dynSystem} is a quadratic function of input $\uv$ for all states outside the origin, $\x \in \X \setminus \{\myvec{0}\}$. When $\myvec{\Theta}$ is negative definite, the action-value function $Q$ is concave and has a unique maximum. 
	\end{proposition}
	
	The proof is Appendix \ref{sec:app:1}.
	
	Next, we present finding the maximum on i$^{th}$ axis using least squares linear regression with polynomial features. 
	\begin{definition}
		\textit{$Q$-axial restriction} on $i^{\text{th}}$ axis is a univariate function 
		$\qifunctionZ{i} = Q(\x, u\myvec{e_i})$, s.t. $\myvec{e_i}$ is a unit vector on $i^{th}$ axis.
	\end{definition}
	
	$Q$-axial restriction on $i^{\text{th}}$ axis is a quadratic function,
	\begin{equation}
	\label{eq:qxi}
	\qifunctionZ{i} = \qcoefsi^T [u^2 \; u \; 1]^T,
	\end{equation}
	for some vector $\qcoefsi = [p_{2,i} \; p_{1,i}\; p_{0,i}]^T \in \R{3}$ based on Proposition \ref{th:prop}. Our goal is to find $\qcoefsi$ by sampling the input space $U$ at a fixed state.

	Suppose, we collect $d_n$ input samples in the $i^{th}$ axis,
	$U_i = [\usample{1} \; ... \; \usample{d_n}]^T.$
	The simulator returns state outcomes when the input samples are applied to the fixed state $\x$, 
	$ X_i = [\x'_{1,i}\;...\;\x'_{d_n,i}]^T,$ where 
	$\x'_{j,i} \leftarrow D(\x, \usample{j}), \;\; j=1,...,d_n. $
	Next, $Q$-estimates are calculated with \eqref{eq:q},
	\begin{equation}
	\myvec{Q_i} = [\qsampleZ{1}\;...\;\qsampleZ{d_n}]^T,
	\label{eq:qi}
	\end{equation}
	where 
	$\qsampleZ{j} = \theta^T F(\x'_{j,i}), \;\; j=1,...,d_n. 
	$ Using the supervised learning terminology the Q estimates, $\myvec{Q_i}$, are the labels that match the training samples $U_i$.
	Matrix,
	\begin{equation}
	C_i = \begin{bmatrix}
	(\usample{1})^2 & \usample{1} & 1 \\ 
	(\usample{2})^2 & \usample{2} & 1 \\ 
	&...  & \\ 
	(\usample{d_n})^2 & \usample{d_n} & 1 
	\end{bmatrix},
	\label{eq:ci}
	\end{equation}
	contains the training data projected onto the quadratic polynomial space. The solution to the supervised machine learning problem,
	\begin{equation}
	\label{eq:reg}
	C_i \qcoefsi = \myvec{Q_i}
	\end{equation}
	fits $\qcoefsi$ into the training data $C_i$ and labels $\myvec{Q_i}$.
	The solution to \eqref{eq:reg},
	\begin{equation}
	\label{eq:regRes}
	\hat{\qcoefsi} = \argmin_{\qcoefsi} \sum_{j=1}^{d_n} (C_{j,i} \qcoefsi - \qsampleZ{j})^2
	\end{equation}
	is the coefficient estimate of the $Q$-axial restriction \eqref{eq:qxi}. A solution to $\frac{d\qifunctionZ{i}}{du} = 0$ is a critical point, and because $Q$ is quadratic the critical point is
	\begin{equation}
	\label{eq:hatu}
	\hat{u}^*_i = -\frac{\hat{p}_{1,i}}{2\hat{p}_{2,i}}.
	\end{equation}
	Lastly, we ensure that action selection falls within the allowed action limits,
	\begin{equation}
	\label{eq:max}
	\hu{i} = \min ( \max (\hat{u}^*_i, u^l_i), u^u_i),
	\end{equation}
	where $u^l$ and $u^u$ are lower and upper acceleration bounds on the $i^{th}$ axis, respectively.
	
	Repeating the process of estimating the maxima on all axes and obtaining $\huv{i} = [\hu{1},...,\hu{\na}]$, we calculate the final policy with
	\begin{equation}
	\label{eq:lsaspa}
	\policyApp = 
	\left\{\begin{matrix}
	\myvec{\pi_c}^{Q}(\x), & Q(\x, \myvec{\pi_c}^{Q}(\x)) \geq Q(\x, \myvec{\pi_n}^{Q}(\x)) \\ 
	\\
	\myvec{\pi_n}^{Q}(\x), & \text{otherwise}
	\end{matrix}\right.
	\end{equation}
	where
	\begin{align*}
		\myvec{\pi_c}^{Q}(\x) &= \na^{-1}\myvec{\pi_n}^{Q}(\x) & \text{(convex policy)}\\
		\myvec{\pi_n}^{Q}(\x) &=  \sum_{i=1}^{\na}  \hu{i} \ev{i},  & \text{(non-convex policy)}
	\end{align*}
	
	The policy approximation \eqref{eq:lsaspa} combines the vector sum of the non-convex policies
	\eqref{eq:max} with the convex sum policy. The convex sum
	guarantees the system's monotonic progression towards the goal for a deterministic system \cite{faust-acta-13}, but the simple
	vector sum (non-convex policy) does not \cite{faust-acta-13}. If, however, the vector sum
	performs better than the convex sum policy, then \eqref{eq:lsaspa} allows us
	to use the better result.
	The disturbance changes the $\myvec{Q}$ function, and the regression 
	fits Q to the observed data. 
	
	\section{Case studies}
	\label{sec:results}
	Table \ref{tab:app} summarizes the case studies used for evaluation.  
	\frm\ was implemented in MATLAB 2013a and 2014a and all training and simulations are executed on either a single core of Intel i3-2120 at 3.3GHz with 4GB RAM or Intel Xeon E5-1620 at 3.6GHz with 8GB RAM. Experiments were performed on an AscTec Hummingbird Quadrotor equipped with a \dist{0.62} suspended load weighing 45 grams. The quadrotor and load positions were captured with a motion capture system at 100 Hz. The testbed's internal tracking system tracked the quadrotor position using a LSAPA generated trajectory as a reference. The video\footnote{\url{https://youtu.be/ZkFt1uY6vlw}} contains the experiments and visualization of the simulations.
	
	\begin{table*}
		\centering
		\caption{\small Summary of case studies. The table shows the task, Section that discusses the case study, and for both PEARL phases, training and planning, and state space dimensionality, state space volume, and task variant. The table also shows the magnitude of the task complexity between training and Planning.}
		\label{tab:app}
		\centering
        \begin{tabular} {l|l|l|r|r|l}
			\hline\noalign{\smallskip}
			Task & Location & PEARL & State space & State space & Task variation\\
            &  & phase & dimensionality &  volume & \\\hline
			Multi agent pursuit & Section~\ref{sec:res:multi} & Training & 12 & $0.8^{12}$ & Static prey; 3 pursuit agents \\
                                                           &  & Planning & 50 & $5.0^{50}$ & Moving prey, 25-agent agents in real-time, 1000 offline \\
             &  & Magnitude & 8.33 times & ~$10^{36}$ times &  \\\hline
			Dynamical obstacle & Section~\ref{sec:res:obs} & Training & 4 & $50\,\pi$ & 4 static obstacles \\
            avoidance                                    &  & Planning & 4 & $1\,250\,\pi$ & Up to 900 stochastically moving obstacles \\
                                                                  &  & Magnitude & Unchanged & $25$ times & \\\hline
			Aerial cargo delivery & Section~\ref{sec:res:suspended} & Training & 10 & $1.8\,10^{4}$ & Without disturbance \\
                                                                 &  & Planning  & 10 & ~$6.1\,10^{8}$ & Stochastic disturbance \\
                                                                 &  & Magnitude & Unchanged & $3.4\,10^{4}$ times &  \\\hline
			Rendezvous & Section~\ref{sec:res:rendezvous} & Training & 16 & $4.3\,10^3$ & Without disturbance \\
             										  &  & Planning & 16 & $1.5\,10^{10}$ & Stochastic disturbance \\
                                                      &  & Magnitude & Unchanged & ~$3.4\, 10^{6}$  times &  \\\hline
			Flying inverted & Section~\ref{sec:res:flyinv} & Training & 10 & $4.8^3$ & Without disturbance \\
             pendulum &  & Planning & 10 & $4.8^3$  & Stochastic disturbance \\
             &  & Magnitude & Unchanged & Unchanged & \\\hline
             
            \end{tabular}
	\end{table*}
	
	\subsection{Multi-agent pursuit}
	\label{sec:res:multi}
	Variants of  multi-agent pursuit tasks have been studied in the RL and multi-agent literature \cite{nn-pursuit,nowe-rlartMulti-12,wu-largescale-10,boids-87}. A typical predator-pursuit task works in a grid world with multiple-agents pursing a prey \cite{nn-pursuit}. In the classical variant, both pursuers and the prey are autonomous agents \cite{wu-largescale-10}. The pursuers work together as a team of independent agents \cite{wu-largescale-10}. Ogren et al.
	\cite{ogren-lypunov-02} approaches multi-agent coordination through the construction of a control Lyapunov function that defines a desired formation, similar to our proposed solution. But, their approach uses a given control law to move the agents, while we use a RL-derived planner. The swarming community addresses a similar problem of a large group of agents working together with a limited information exchange \cite{nowe-rlartMulti-12}. Typically, the resulting complex system exhibits an emergent behavior as a result of only few given rules. For example, Boids are intended to model the behavior of bird flocks, and accomplish that using a sum of separation, coherence, and alignment proportional controllers for each agent \cite{boids-87}. In other work, Kolling et al. compute synchronized trajectories for UAVs that guarantee the detection of all targets \cite{kolling-uav-aamas13}. 
	
	Our variant of the multi-agent pursuit task controls only the pursuers chasing the prey, while the prey follows a predetermined trajectory unknown to the pursuers. The problem formulation we use differs from the typical formulation in three ways. First, it works in an arbitrarily large continuous state and action space rather than the discretized grid world. Second, it learns and plans in joint agent-position-velocity space, and handles variable team size without additional learning. Lastly, we require the pursuers to keep distance among themselves, thus we learn and plan in the joint pursuers' space. We formulate the task as follows:
	\begin{mydef} \textit{(Multi-agent pursuit task.)}
		A set of $r$ agents must follow the prey agent in close proximity, while maintaining separation to each other.  
		The agents know the current position and velocity of the prey (leader) but do not know the future headings.
	\end{mydef}
	
    \subsubsection{PEARL setup}
	To solve this task, we set up a MDP and preferences as outlined in Section \ref{sec:feature}. Assuming $\nr$ agents, with $\nr_i$  DoFs each, the state space is the joint agent-position-velocity vector space $\X \subset \R{2 \nx}$. State $\x$'s coordinates, $\x_{ij}, \dx_{ij},$ $i=1,..,\nr,$ $j=1..\nr_i$, denote $i^{th}$'s agent position and velocity in the direction of axis $j$. The action space consists of acceleration vectors, $\U \subset\R{\na}$, where $\na = \sum_{i=i}^{\nr}{\na}_i$ and $\ddx_{ij}$ is $i^{th}$ agent's acceleration in the $j$ direction.
	To design the preferences for the pursuit task, we look at the task definition. There are three preferences: close proximity to the prey (attractor), following the prey (attractor), and maintaining distance between the agents (repeller). 
	Thus, the feature vector has three components, $$\F(\x,\nx) = [\Fi_1(\x,\nx)\;\; \Fi_2(\x,\nx)\;\; \Fi_3(\x,\nx)]^T.$$ 
	Following the method from Section \ref{sec:feature}, we express the distance to the prey as $$\Fi_1(\x,\nx) = \sum_{i=1}^{\nr} \sum_{j=1}^{\nr_i}(\x_{ij} - \myvec {p}_j)^2,$$ 
	and  \textit{following the prey}, as reducing the difference in velocities between the agents and the prey, $$\Fi_2(\x,\nx) = \sum_{i=1}^{\nr} \sum_{j=1}^{\nr_i} ({\dx}_{ij} - \dot{\myvec{p}}_j)^2$$ where $\myvec{p}_i$ 
	and $\dot{\myvec{p}_i}$ are prey's position and velocity in the direction $i$. The last feature is increasing the distance between the agents,
	$$\Fi_3(\x,\nx)= (1+\sum_{i,j=1}^{\nr} \sum_{k=1}^{\nr_i} (\x_{ik}- \x_{jk})^2)^{-1}.$$ 
	
	\subsubsection{Learning} To learn the pursuit task, we use 3 planar agents with double integrator dynamics ($\nr_i=2, \,i=1,2,3$, $\nr=3$). The pursuit team of three robots is the smallest team for which all features are non-trivial. The maximum absolute accelerations less than \accel{3}. We run CAFVI \cite{faust-acta-13} 
	as the learning agent in \frm\ with DAS \cite{faust-acta-13}  for 300 iterations to learn the feature vector weights. The sampling space is inside a six-dimensional hypercube $[\dist{-0.4},\, \dist{0.4}]^{12}$. The prey is stationary at the origin during learning. The resulting weights are $\teta^T = [-16.43 -102.89 -0.77]^T$. Simulations run at 50 Hz. The time to learn is \dur{145}.
	
	\subsubsection{Planning} To plan a task, we assign a trajectory to the prey, and increase the number of agents. The prey starts at the origin, and the pursuers start at random locations within \dist{5} from the origin. We plan a \dur{20} trajectory. 
	
	Figure  \ref{fig:warf14_flocking_timing} depicts the planning computational time for a \dur{20} trajectory as the number of pursuers increases. State and action spaces grow with the team size, and the planning time stays polynomial. This is because the feature vector scales polynomially with the state size. 
	The method plans the problems with continuous state spaces up to $\R{100}$ and continuous actions up to $\R{50}$ in real-time  (below the green line in Figure  \ref{fig:warf14_flocking_timing}), as trajectory computation takes less time than its execution. 
	
    \subsubsection{Results}
	
	\begin{figure}[tb]
		\centering
		\includegraphics[width=0.40\textwidth,trim=0mm 0mm 0mm 10mm,clip]{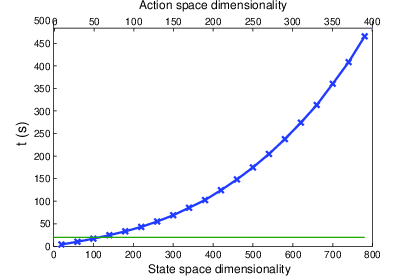}
		\caption{Planning time for a 20 second trajectory with increasing state space dimensionality. Recall for this multi-agent pursuit problem, the state space is two times the action space size.  Problems below the solid green line can be solved in real-time.}
		\label{fig:warf14_flocking_timing}       
	\end{figure}

	\begin{figure*}[tb]
		\begin{center}
			\begin{tabular}{ccc}
				\subfloat[\small Spiral - xy view]{\includegraphics[trim=0mm 0mm 5mm 3mm, clip,width=0.30\textwidth,height=2.5cm,keepaspectratio=false]{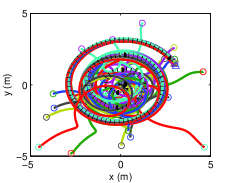}\label{fig:wafr_flocking_agent25_Spiral_xy}} &
				\subfloat[\small Spiral - x view]{\includegraphics[trim=0mm 0mm 5mm 3mm, clip,width=0.30\textwidth,height=2.5cm,keepaspectratio=false]{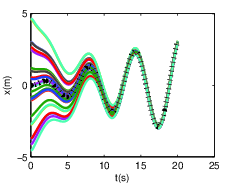}\label{fig:wafr_flocking_agent25_Spiral_x}} &
				\subfloat[\small Spiral - y view]{\includegraphics[trim=0mm 0mm 5mm 3mm, clip,width=0.30\textwidth,height=2.5cm,keepaspectratio=false]{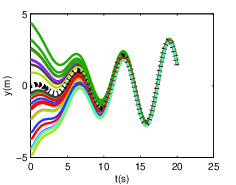}\label{fig:wafr_flocking_agent25_Spiral_y}}\\
				\subfloat[\small Lemniscate - xy view]{\includegraphics[trim=0mm 0mm 5mm 3mm, clip,width=0.30\textwidth,height=2.5cm,keepaspectratio=false]{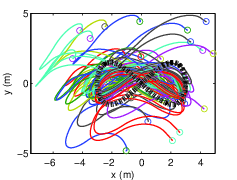}\label{fig:wafr_flocking_agent25_Eight_xy}} &
				\subfloat[\small Lemniscate - x view]{\includegraphics[trim=0mm 0mm 5mm 3mm, clip,width=0.30\textwidth,height=2.5cm,keepaspectratio=false]{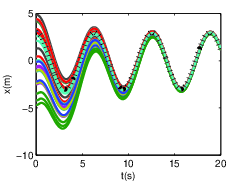}\label{fig:wafr_flocking_agent25_Eight_x}} &
				\subfloat[\small Lemniscate - y view]{\includegraphics[trim=0mm 0mm 5mm 3mm, clip,width=0.30\textwidth,height=2.5cm,keepaspectratio=false]{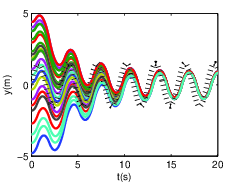}\label{fig:wafr_flocking_agent25_Eight_y}}\\ 
				
				\subfloat[\small Brownian vel. - xy view]{\includegraphics[trim=3mm 0mm 5mm 3mm, clip,width=0.30\textwidth,height=2.5cm,keepaspectratio=false]{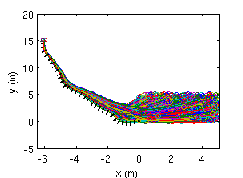}\label{fig:wafr_flocking_agent1000_Brown_xy}} &
				\subfloat[\small Brownian vel. - x view]{\includegraphics[trim=3mm 0mm 5mm 3mm, clip,width=0.30\textwidth,height=2.5cm,keepaspectratio=false]{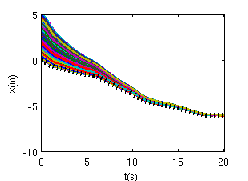}\label{fig:wafr_flocking_agent1000_Brown_x}} &
				\subfloat[\small Brownian vel. - y view]{\includegraphics[trim=3mm 0mm 5mm 3mm, clip,width=0.30\textwidth,height=2.5cm,keepaspectratio=false]{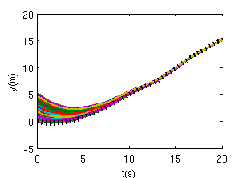}\label{fig:wafr_flocking_agent1000_Brown_y}}\\
			\end{tabular}\\	
			\caption{\small Multi-agent pursuit task learning transfer. Three different pursuit tasks planned with the same learning. The prey following a spiral (a-c) and a lemniscate curve (d-f) chased by 25 agents. 1000-agent pursuit task of a prey that  follows a random acceleration trajectory (g-i). The prey's trajectory is a dotted black line. 
				\label{fig:wafr_flocking_agent25_Spiral}}
			\label{fig:wafr_flocking_agent25_Eight}
			\label{fig:wafr_flocking_agent1000_Brown}
		\end{center}
	\end{figure*}

	\begin{figure*}[tb]
		\begin{center}
			\begin{tabular}{ccc}
				\subfloat[Spiral - xy view]{\includegraphics[trim=0mm 0mm 0mm 0mm, clip,width=0.3\textwidth,height=2.5cm,keepaspectratio=false]{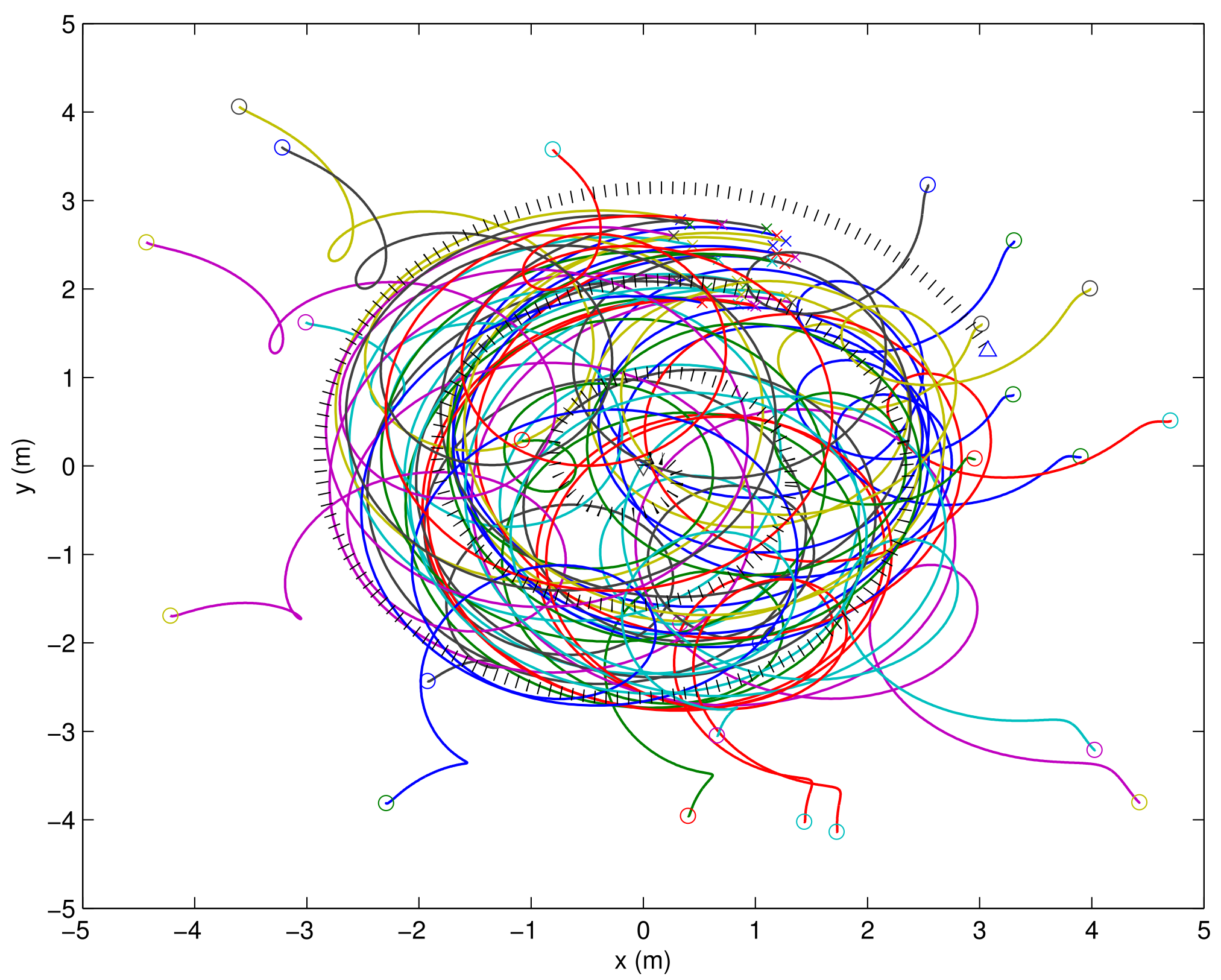}\label{fig:wafr_flocking_agent25_Spiral_xy_boids}} &
				\subfloat[Spiral - x view]{\includegraphics[trim=0mm 0mm 0mm 0mm, clip,width=0.3\textwidth,height=2.5cm,keepaspectratio=false]{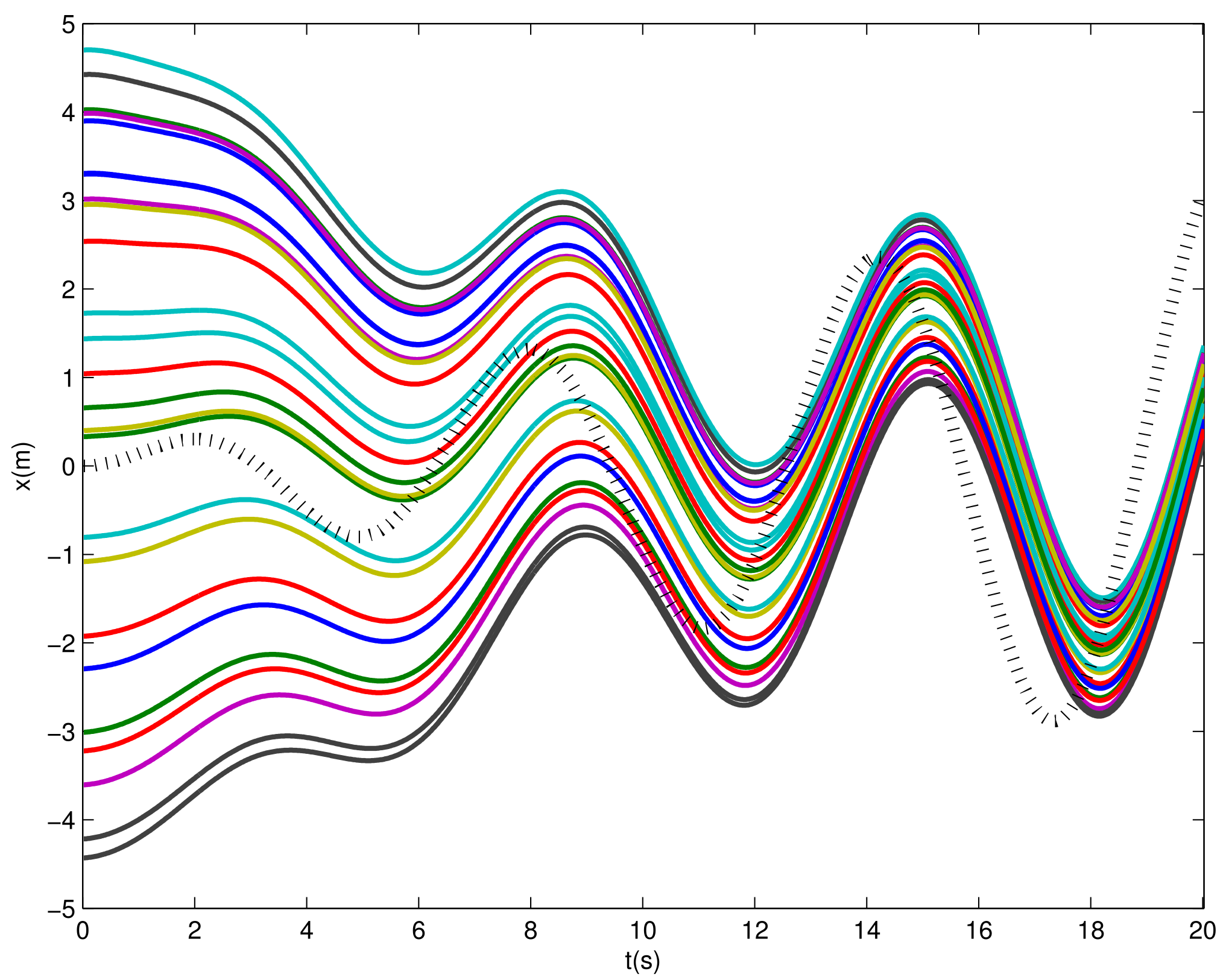}\label{fig:wafr_flocking_agent25_Spiral_x_boids}} &
				\subfloat[Spiral - y view]{\includegraphics[trim=0mm 0mm 0mm 0mm, clip,width=0.3\textwidth,height=2.5cm,keepaspectratio=false]{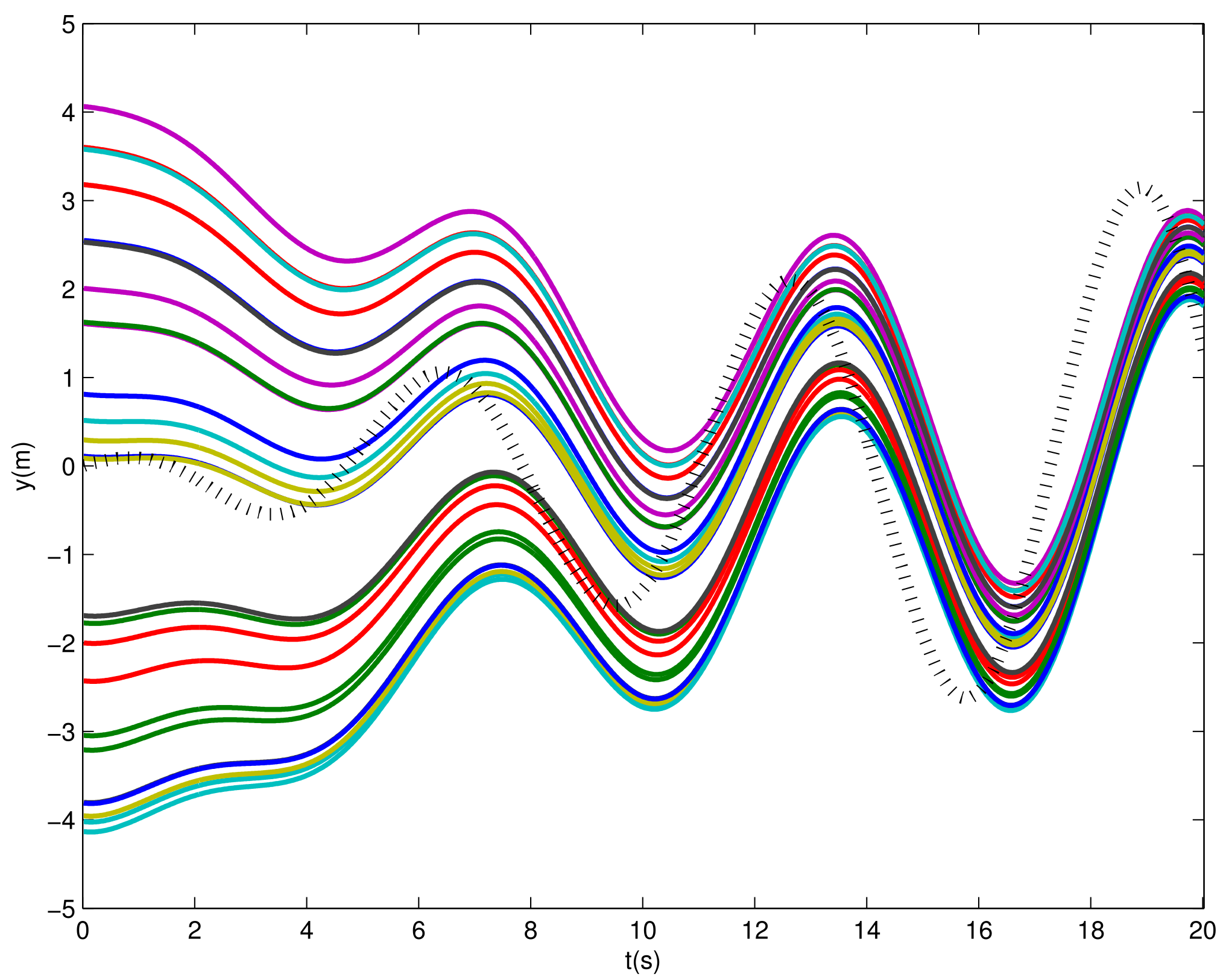}\label{fig:wafr_flocking_agent25_Spiral_y_boids}}\\
				\subfloat[xy view]{\includegraphics[trim=0mm 0mm 0mm 0mm, clip,width=0.3\textwidth,height=2.5cm,keepaspectratio=false]{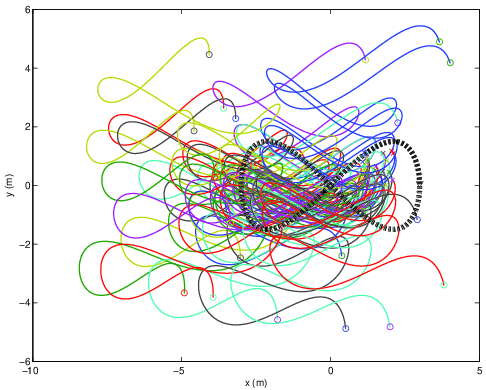}\label{fig:wafr_flocking_agent25_Eight_xy_boids}} &
				\subfloat[Lemniscate - x view]{\includegraphics[trim=0mm 0mm 0mm 0mm, clip,width=0.3\textwidth,height=2.5cm,keepaspectratio=false]{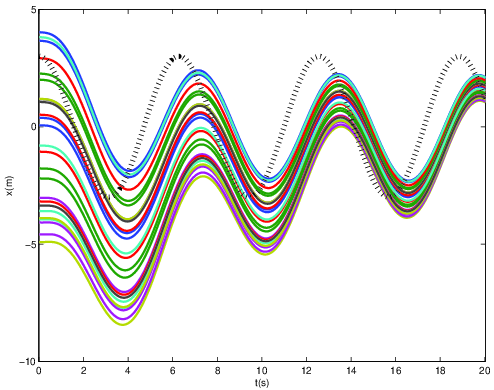}\label{fig:wafr_flocking_agent25_Eight_x_boids}} &
				\subfloat[Lemniscate - y view]{\includegraphics[trim=0mm 0mm 0mm 0mm, clip,width=0.3\textwidth,height=2.5cm,keepaspectratio=false]{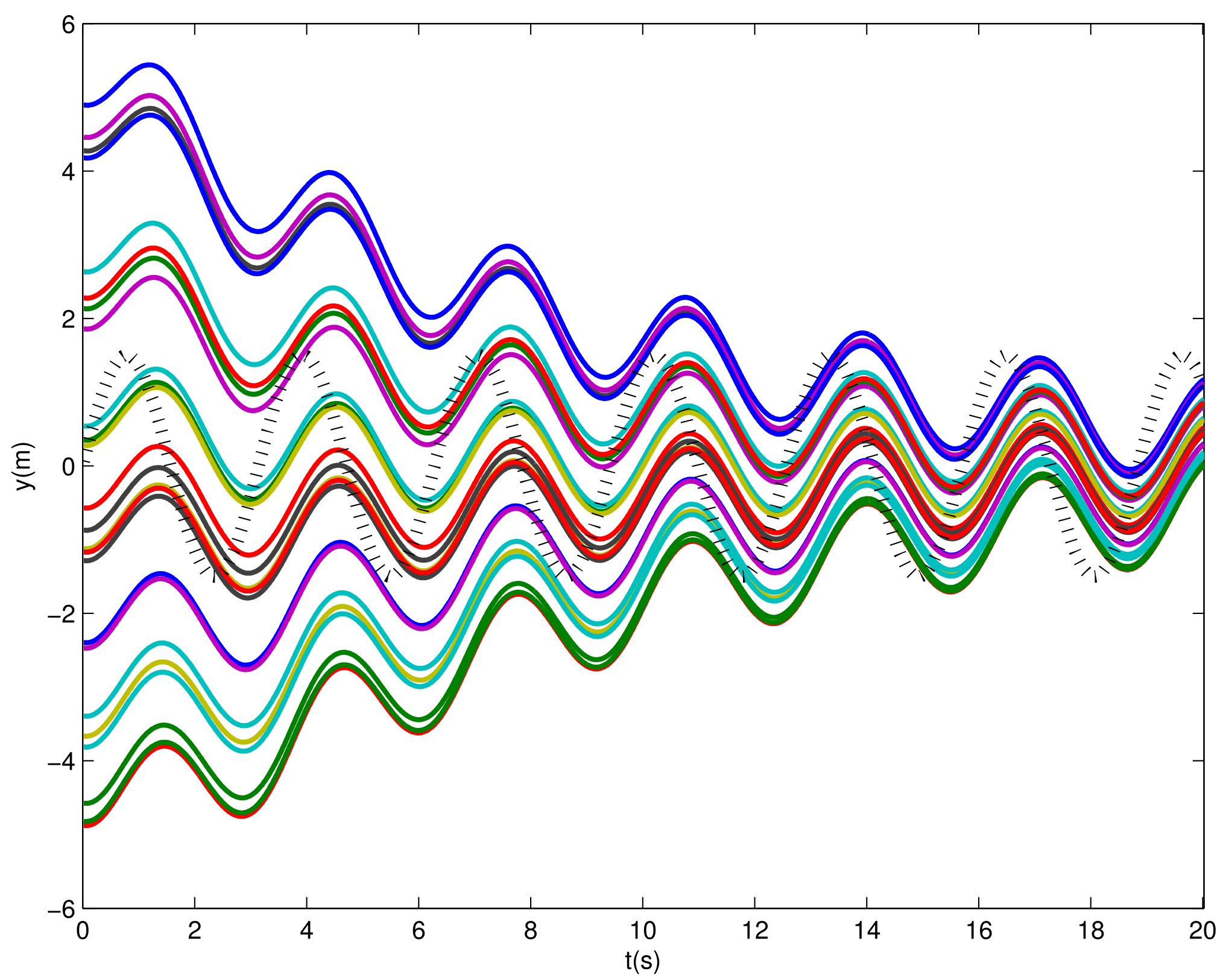}\label{fig:wafr_flocking_agent25_Eight_y_boids}}\\ 
				
				\subfloat[Brownian velocity - xy view]{\includegraphics[trim=0mm 0mm 0mm 0mm, clip,width=0.3\textwidth,height=2.5cm,keepaspectratio=false]{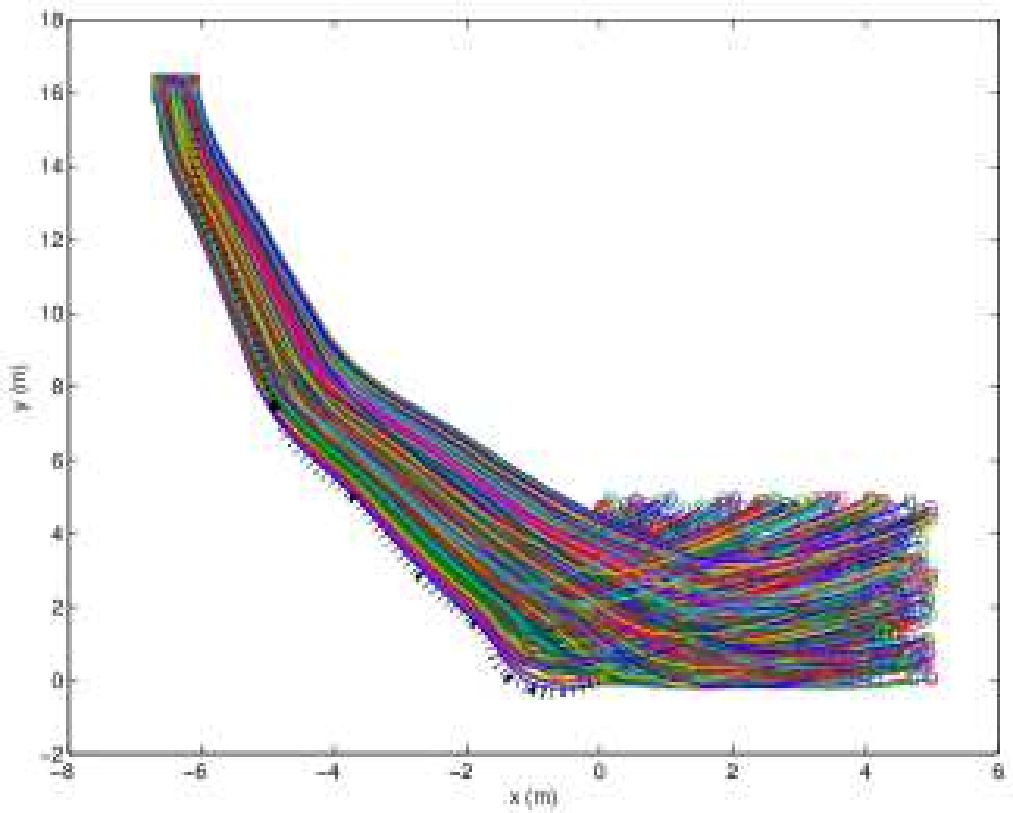}\label{fig:wafr_flocking_agent1000_Brown_xy_boids}} &
				\subfloat[Brownian velocity - x view]{\includegraphics[trim=0mm 0mm 0mm 0mm, clip,width=0.3\textwidth,height=2.5cm,keepaspectratio=false]{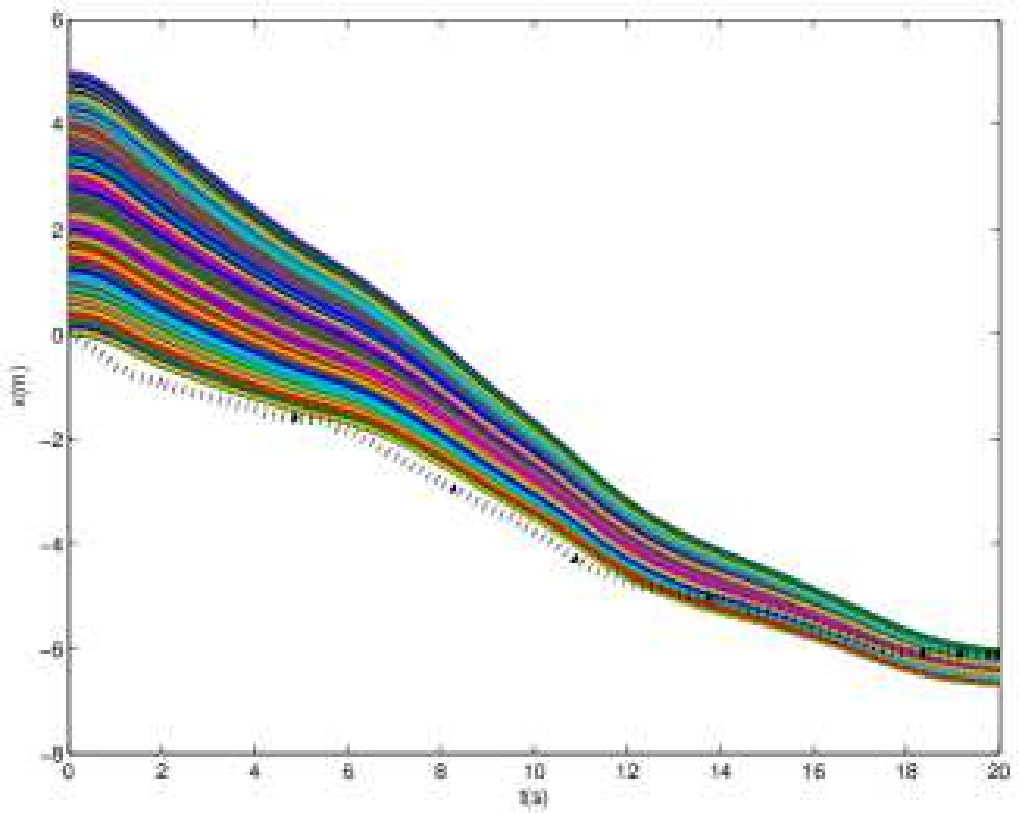}\label{fig:wafr_flocking_agent1000_Brown_x_boids}} &
				\subfloat[Brownian velocity- y view]{\includegraphics[trim=0mm 0mm 0mm 0mm, clip,width=0.3\textwidth,height=2.5cm,keepaspectratio=false]{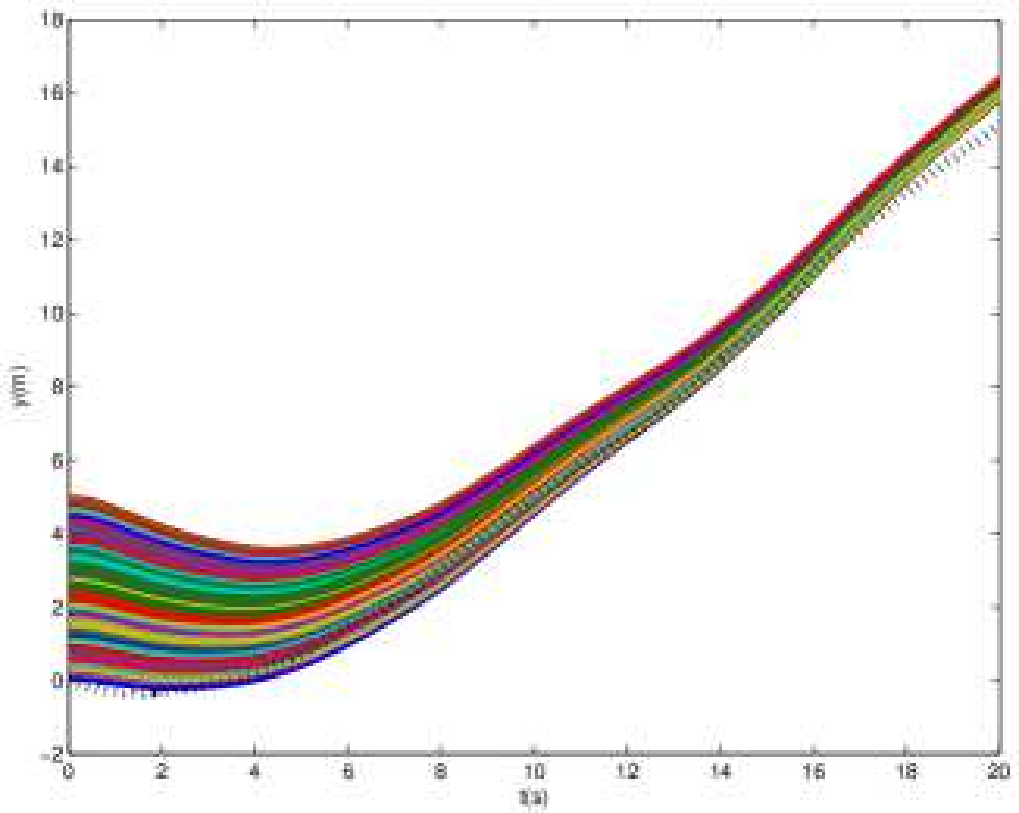}\label{fig:wafr_flocking_agent1000_Brown_y_boids}}\\
			\end{tabular}\\	
			\caption{\small Three different pursuit tasks planned with Boids. The prey following a spiral (a-c) and a lemniscate curve (d-f) chased by 25 agents. 1000-agent pursuit task of a prey that  follows a random acceleration trajectory (g-i). The prey's trajectory is a dotted black line. 
				\label{fig:wafr_flocking_agent25_Spiral_boids}}
			\label{fig:wafr_flocking_agent25_Eight_boids}
			\label{fig:wafr_flocking_agent1000_Brown_boids}
		\end{center}
	\end{figure*}

	The evaluation looks at 25-agent pursuit of a prey that tracks either a spiral (Figures \ref{fig:wafr_flocking_agent25_Spiral_xy} - \ref{fig:wafr_flocking_agent25_Spiral_y}) or lemniscate de Gerono curve (Figures \ref{fig:wafr_flocking_agent25_Eight_xy}-
	\ref{fig:wafr_flocking_agent25_Eight_y}). The agents start at random locations uniformly drawn within \dist{5} from the prey's initial position (origin). Figures \ref{fig:wafr_flocking_agent25_Spiral_xy} and  \ref{fig:wafr_flocking_agent25_Eight_xy} show the movement of the agents in the xy-plane. Although the pursuers do not know the future prey's positions, they start tracking in increasingly close formation. Figures \ref{fig:wafr_flocking_agent25_Spiral_x} and  \ref{fig:wafr_flocking_agent25_Eight_x} show the x-axis trajectory over time, and Figures \ref{fig:wafr_flocking_agent25_Spiral_y} and  \ref{fig:wafr_flocking_agent25_Eight_y} show the y-axis trajectory over time. Note that the initial positions are uniformly distributed, and after \dur{7} all agents are close to the prey and remain in phase with it. The only exception is the y-coordinate of the lemniscate (Figure  \ref{fig:wafr_flocking_agent25_Eight_y}). Here, the agents never catch up with the prey. The higher frequency oscillations in the y-direction make the prey hard to follow, but the agents maintain a constant distance from it.
	Figures \ref{fig:wafr_flocking_agent1000_Brown_xy} -
	\ref{fig:wafr_flocking_agent1000_Brown_y} depict a \dur{20} trajectory of a 1000-pursuer task. The state space is $\R{4000}$ and the action space is $\R{2000}$. DAS successfully plans a trajectory in this large-scale space. The prey's trajectory has Brownian velocity (its acceleration is randomly drawn from a normal distribution). 
	In this example, the prey travels further from the origin than the previous two examples.
	
	\begin{table*}[tb]
		\centering
		\caption{\small Trajectory characteristics after 20 seconds of simulation as a function of the prey's trajectory and the number of concurrent agents. State (\# State) and action (\# Action) space dimensionality, computational time (Comp.), average distance from the prey (prey Dist.), and distance between the agents (Dist. Agents) are shown. All results are averaged over 100 trajectories starting from randomly drawn initial states within 5 meters from the origin. }
		\label{tab:flocking}       
		\begin{tabular} {l|l|r|r|r|r|rr|rr}
			\hline\noalign{\smallskip}
			Method&Prey & \# & \multicolumn{1}{p{1cm}|}{\# State}& \multicolumn{1}{p{1.5cm}|}{\# Action} & Comp. & \multicolumn{2}{p{2.5cm}|}{Prey Dist. (m)} &  \multicolumn{2}{p{2.5cm}}{Dist. Agents $(m)$} \\\hline
			\multirow{15}{*}{PEARL}
			&& 5 & 20& 10& 4.18& 0.29 & 0.01 & 0.46 & 0.00 \\ 
			&& 10 & 40& 20& 7.26& 0.16 & 0.00 & 0.23 & 0.00 \\ 
			& Line& 15 & 60& 30& 10.62& 0.11 & 0.00 & 0.16 & 0.00 \\ 
			&& 20 & 80& 40& 13.62& 0.09 & 0.00 & 0.12 & 0.00 \\ 
			&& 25 & 100& 50& 17.21& 0.08 & 0.00 & 0.11 & 0.00 \\ 
			\noalign{\smallskip}\cline{2-10}\noalign{\smallskip}
			&& 5 & 20& 10& 4.05& 0.34 & 0.01 & 0.46 & 0.00 \\ 
			&& 10 & 40& 20& 7.22& 0.24 & 0.01 & 0.23 & 0.00 \\ 
			&Spiral& 15 & 60& 30& 10.21& 0.22 & 0.01 & 0.15 & 0.00 \\ 
			&& 20 & 80& 40& 13.31& 0.22 & 0.01 & 0.12 & 0.00 \\ 
			&& 25 & 100& 50& 16.98& 0.22 & 0.01 & 0.10 & 0.00 \\ 
			\cline{2-10}\noalign{\smallskip}
			&& 5 & 20& 10& 4.08& 0.36 & 0.01 & 0.46 & 0.00 \\ 
			&& 10 & 40& 20& 7.18& 0.29 & 0.01 & 0.22 & 0.00 \\ 
			&Lemniscate& 15 & 60& 30& 10.21& 0.27 & 0.01 & 0.15 & 0.00 \\ 
			&& 20 & 80& 40& 13.33& 0.26 & 0.00 & 0.11 & 0.00 \\ 
			&& 25 & 100& 50& 16.90& 0.26 & 0.00 & 0.09 & 0.00 \\ 
			\noalign{\smallskip}\hline\noalign{\smallskip}
			
			\multirow{15}{*}{Boids}
			&& 5 & 20& 10& 0.82&  0.83 & 0.17 & 0.60 & 0.12 \\ 
			&& 10 & 40& 20& 0.82&  0.85 & 0.13 & 0.61 & 0.07 \\ 
			&Line& 15 & 60& 30& 0.82&  0.85 & 0.10 & 0.62 & 0.05 \\ 
			&& 20 & 80& 40& 0.83&  0.83 & 0.09 & 0.62 & 0.05 \\ 
			&& 25 & 100& 50& 0.84& 0.84 & 0.08 & 0.62 & 0.05 \\ 
			\cline{2-10}\noalign{\smallskip}
			&& 5 & 20& 10& 0.82& 2.53 & 0.18 & 0.61 & 0.12 \\ 
			&& 10 & 40& 20& 0.83& 2.55 & 0.14 & 0.63 & 0.07 \\ 
			&Spiral& 15 & 60& 30& 0.85& 2.55 & 0.10 & 0.62 & 0.06 \\ 
			&& 20 & 80& 40& 0.85& 2.52 & 0.10 & 0.62 & 0.05 \\ 
			&& 25 & 100& 50& 0.85& 2.53 & 0.08 & 0.62 & 0.04 \\ 
			\cline{2-10}\noalign{\smallskip}
			&& 5 & 20& 10& 0.84& 0.77 & 0.16 & 0.59 & 0.11 \\ 
			&& 10 & 40& 20& 0.83& 0.81 & 0.13 & 0.62 & 0.07 \\ 
			&Lemniscate& 15 & 60& 30& 0.84& 0.78 & 0.09 & 0.62 & 0.06 \\ 
			&& 20 & 80& 40& 0.87& 0.77 & 0.10 & 0.62 & 0.05 \\ 
			&& 25 & 100& 50& 0.86& 0.76 & 0.08 & 0.63 & 0.04 \\\hline
		\end{tabular}
	\end{table*}

	We compare the PEARL multi-agent pursuit task with Boids \cite{boids-87}. Boids uses the three standard goals; separation, alignment, and cohesion (independent of the prey), and two additional prey-direct rules; prey cohesion, and prey alignment. Since our agents have a double integrator dynamics, the alignment and cohesion rules are realized by minimizing the distance and velocity differential. We empirically tuned the rule weights until Boids started pursuing the prey. The resulting weights are $1,\, 0.01,\, 0.01, \,0.1,\, 1$ for flock separation, alignment, and cohesion and for prey alignment and cohesion, respectively. Figures \ref{fig:wafr_flocking_agent1000_Brown_xy_boids} -
	\ref{fig:wafr_flocking_agent1000_Brown_y_boids} show the resulting agents' trajectories. Compared to PEARL trajectories in Figure \ref{fig:wafr_flocking_agent25_Spiral_boids}, Boids agents lag behind more in the Spiral and Lemniscate trajectories. They also maintain a larger distance among agents, which is expected given the explicit separation rule that comes to effect when two agents are less than \distcm{10} apart.
	
	Table \ref{tab:flocking} shows the summary of ending trajectory characteristics averaged over 100 trials with the prey following the given trajectories.
	For each trajectory we plan the pursuit task with a varying number of pursuers. For all tasks the \dur{20} pursuit takes less than \dur{20} to compute even in the 100-dimensional state spaces (Table \ref{tab:flocking}), confirming the timing benchmark in Figure  \ref{fig:warf14_flocking_timing}. The next column, \textit{prey distance}, shows that the average pursuer-prey distance remains below \distcm{30}. The minimal standard deviation suggests consistent results, in agreement with the trajectory plots (Figure  \ref{fig:wafr_flocking_agent25_Spiral}). The last column shows the average distance between the agents that was to be maximized. The distance ranges between \distcm{11} and \distcm{46}. It decreases as the pursuer team grows, reflecting higher agent density around the prey and a tighter formation. 
	Motion animation shows that while approaching the prey, the agents maintain the same formation and relative positioning, thus avoiding collisions. Compared to \frm, Boids are an order of magnitude faster generating trajectories, but the average distance to the prey is an order of magnitude larger.

	
	\subsection{Dynamic obstacle avoidance}
	\label{sec:res:obs}
	
	Planning motion in dynamic environments is challenging because plans must be frequently adjusted due to moving obstacles.  To address this challenge, proposed planners vary greatly in methodology, information about the position and dynamics of the obstacles, and the ability to account for stochastic obstacle motion \cite{chiang2017safety}.   
	For example, APF methods provide fast solutions by using only the positional information of obstacles near to the robot \cite{lccf-hcrntahhrc-11,ge2002dynamic}. The challenge of APF methods lies in tuning the  repulsive and attractive potentials. 
	On the other hand, state of the art sampling-based methods for dynamic environments plan in state-time space in order avoid guiding the robot into inevitable collision states, states which lead to collision regardless of control policies \cite{benenson2006integrating,chiang2017dynamic}.     These methods often require knowledge of the dynamics of obstacles (deterministic or stochastic), which can be difficult to obtain \cite{chiang2017safety}. A method that approximates an obstacle's dynamics is Velocity Obstacle (VO) that computes control actions for collision avoidance in the robot’s velocity space using the geometry \cite{fiorini1998motion}.   In VO the obstacles are approximated as circles or spheres and are assumed to move at a constant velocity, which can be inaccurate in crowded environments with stochastically moving obstacles. In this study, we define the dynamic obstacle avoidance task as follows:
	
	\begin{mydef} \textit{(Dynamic obstacle avoidance task.)}
		The agent must navigate from start to goal without collision with obstacles with obstacle dynamics unknown to the agent. 
		The agent has access to the current position and velocity information of obstacles.
	\end{mydef}
	
	\subsubsection{PEARL setup} 
	We setup a MDP where the state space is the joint vector of
	robot position and velocity, $\X \subset \R{4},\,\x = [\myvec{x}, \dot{\myvec{x}}]^T \in \X$. The action space is the acceleration
	on each axis with dimension $\U=\R{2}$. 
	This task has two natural preferences: 1) minimize the distance to the goal and
	2) maximize the distance from obstacles. 
	Therefore, PEARL's feature vector is formulated as the combination of two preferences:
	$\F(\x) = [F_1(\x)\;\; F_2(\x)]^T$. 
	The first feature is an attractor towards the goal,
	$F_1(\x) = \| \myvec{x} - \myvec {g} \| ^2$, where $\myvec{x}$ is the position of the agent and $\myvec {g}$ is position of the goal.
	The second feature is a repeller from the closest obstacle, 
	$F_2(\x) = (\beta + d^2)^{-1}$, where $\beta$ is a constant empirically selected to be \dist{0.01} and $d$ is the distance to the closest obstacle.
	
	\subsubsection{Learning} We use 4 stationary obstacles placed at
	$[\dist{3},\,\dist{0}],$ $[\dist{0},\,\dist{3}],$ $[\dist{0},\,\dist{-3}],$
	$[\dist{-3},\,\dist{0}]$ to learn the weights between the two features. 
	The goal is at the origin. The sampling space is inside a two-dimensional
	hypercube $[\dist{-5},\, \dist{5}]^2$. The robot has a maximum speed of
	$\velocity{0.37}$, and a maximum acceleration of $\accel{3}$. 
	We run AVI \cite{ernst-avi} with HOOT policy approximation
	to learn the feature vector weights.  
	The resulting weights are $\teta = [-0.23\;
	-0.1696]^T$. All simulations are done at 10 Hz. The training duration is \dur{123}.
	
	\subsubsection{Planning} 
	The planning environment is illustrated in Figure
	\ref{fig:pathObsRobotAndEnv}. The robot must travel from the start position $[\dist{25},\, \dist{0}]$ to
	the goal at $[\dist{-25},\, \dist{0}]$ under the same speed and acceleration constraints as used for learning. 
	The environment has $N = \{300,$ $450,$ $600,$ $750,$ $900\}$ randomly placed moving obstacles with hybrid stochastic dynamics, three stochastic modes of linear, arc, and swerve. An obstacle is in one of the three modes at any given moment, and the initial mode is randomly determined. 
	Obstacles in linear dynamics mode has a fixed heading but the speed of travel is sampled stochastically from the set $\{ 0.1, 0.2, 0.5, 0.7 \}$ \velocity{} with probability $\{ 0.4, 0.1, 0.2, 0.3 \}$. 
	Obstacles in arc dynamics mode move counter clock-wise with radius \dist{5} at a stochastically sampled angular speed from the set $\{ 0.039, 0.058, 0.088, 0.117 \}$ \si[per-mode=symbol]{\radian \per \second} with probability $\{ 0.4, 0.1, 0.2, 0.3 \}$.
	Obstacles in swerve dynamics mode are changing heading linearly at the rate of $\pi / 3$ \si[per-mode=symbol]{\radian \per \second} between $[-\phi_{invert}, \phi_{invert}]$. $\phi_{invert}$ is sampled uniformly between $[ -\pi / 2, \pi / 2 ]$. 
	All obstacles, regardless of dynamics modes, re-sample stochastically with frequency $\frac{1}{T_{resample}}.$ 
	The obstacles are circles with radius  $r_{obs} = \dist{0.5}$. 
	The average speed of obstacles (\velocity{0.37} regardless of dynamics modes) is identical to the maximum	speed of the robot.
	We maintain the constant density of moving obstacles by restricting the robot
	and moving obstacles to a circle of radius $\dist{50}$. When an
	obstacle hits the boundary of the circle, it is transported to the antipodal
	position on the circle and continues from this new position.
	
	\begin{figure*}[t]
		\begin{center}
			\subfloat[]{\includegraphics[trim=13mm 0mm 29mm 4mm,clip,width=0.3\textwidth,height=4.5cm,keepaspectratio=false]{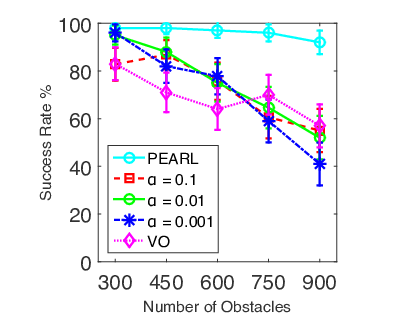}\label{fig:successFig}}
			\subfloat[]{\includegraphics[trim=13mm 0mm 29mm 4mm,clip,width=0.3\textwidth,height=4.5cm,keepaspectratio=false]{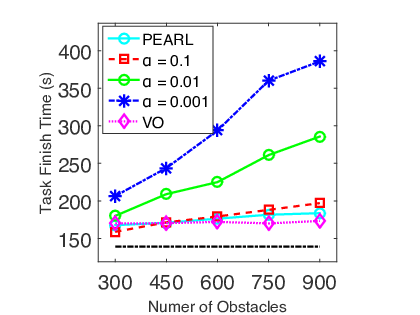}\label{fig:pathLengthFig}}
			\subfloat[]{\includegraphics[trim=13mm 0mm 29mm 4mm,clip,width=0.3\textwidth,height=4.5cm,keepaspectratio=false]{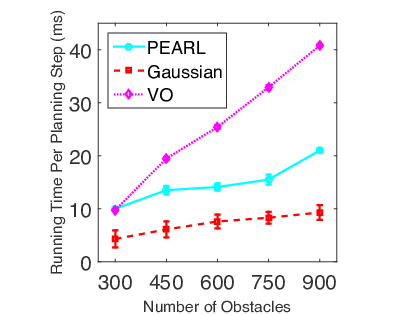}\label{fig:obstimeFig}}
			\caption{\small Planning trajectory characteristics (averaged over 200 trials) for environments with varied complexity (number of obstacles). 
				(a) Task success rate. The error bars are computed using the 99\% confidence interval. (b)
				Amount of time for the agent to reach the goal without collision, the black dotted line is the minimum finish time without obstacles. (c) Computation time per planning step.}
		\end{center}
	\end{figure*}

	\subsubsection{Results}
	Since the agent does not have knowledge of obstacle dynamics in this task, we compared our method with two obstacle avoidance methods that also do not require this knowledge, Gaussian APF and VO. 
	Gaussian APF  considers only the position of obstacles. It combines a quadratic attractive potential toward the goal and a repulsive potential from obstacles
	\cite{lccf-hcrntahhrc-11}. The obstacle potentials are Gaussians with
	$\sigma=\dist{0.45}$ around obstacles, tuned empirically for this problem. The
	relative strength between the attractive and repulsive potential, $\alpha$, has a significant impact on the success rate and also needs to be manually tuned.  Larger $\alpha$ represents a more goal-greedy robot behavior. 
	We compared various values of $\alpha$ to our method. 
	VO considers the position and velocity of
	obstacles \cite{van2011reciprocal} in order to compute velocity obstacles posed by obstacles within \dist{5} of the agent.

	Figure \ref{fig:successFig} and
	\ref{fig:pathLengthFig} show that planning with PEARL has a higher probability of successfully avoiding obstacles, and reaches the goal in less time
	compared to  Gaussian APF. The success rate and task finish time of
	Gaussian APF depends greatly on the parameter $\alpha$. This parameter has to
	be tweaked manually or by optimization algorithms \cite{tw-eapfatairtrpp} after
	many planning trials. PEARL balances the features (similar to finding the
	optimal $\alpha$) in the learning phase with a simplified scenario, and is able
	to transfer the weights to the online plan with comparable or better performance.
	
	Figure \ref{fig:successFig} also shows that
	PEARL has a higher success rate than VO. This is primarily due to
	VO's velocity obstacle formulation that assumes the obstacle has a fixed velocity. 
	This results in trajectories that lead the agent into collision with stochastically moving obstacles.
	PEARL on the other hand, by balancing the features, was able to generate trajectories with sufficient clearance to account for the stochastic obstacle motion.
	In addition, Figure \ref{fig:pathLengthFig} shows the computation time per planning
	step for PEARL is lower than VO, even though PEARL is implemented in MATLAB and VO is in C++. (Our C++ implementation of VO is downloaded from \cite{orca-webpage} and modified to support a single robot and multiple moving obstacles.)   
	PEARL scales linearly with the number of
	obstacles and is capable of generating high success rate trajectories in real-time (Figure
	\ref{fig:obstimeFig}). 
	This suggests that PEARL is a viable
	alternative method for dynamic obstacle avoidance, even when the obstacles are moving with highly unpredictable hybrid stochastic dynamics.

	\subsection{Aerial cargo delivery}
	\label{sec:res:suspended}
	
     We use a quadrotor with a suspended load as our benchmarking platform because it is a popular research platform leading to solutions for multiple robots \cite{sreenath-rss-13}, hybrid systems \cite{cedric}, differentially-flat approaches \cite{sreenath_quad_icra2013}, and load trajectory tracking \cite{palunko-quad-icra-13} among others. The aerial cargo delivery requires a quadrotor, carrying a load on a suspended rigid cable, to deliver the cargo to a given location with the minimal residual load oscillations \cite{faust-quad-icra-13}. The task has applications in delivery supply and aerial transportation in urban environments. The task is easily described, yet, it is difficult for human demonstration as it requires a careful approach to avoid destabilizing the load.
     We use the following definition of the aerial cargo delivery:
	\begin{mydef} \textit{(Aerial cargo delivery task.)}
		The quadrotor equipped with a suspended load must navigate from start to goal arriving at the goal with the minimum residual oscillation of the suspended load. The quadrotor center of the mass is influenced with a stochastic disturbance with Gaussian distribution.
		The agent has access to the quadrotor's position and velocity, and load's position and angular velocity, as well as the location of the goal.
	\end{mydef}
    
	\subsubsection{PEARL setup} The state space is a 10-dimensional joint system, $\x = [\myvec{p}_q, \,\myvec{\eta},\,\dot{\myvec{p}_q},\dot{\myvec{\eta}}]^T$, of the quadrotor's and the load's position and velocity, where $\myvec{p}_q = [x,\,y,\,z]^T$ is the quadrotor's position, and $\myvec{\eta} = [\phi,\,\psi]^T$ is the load's displacement in polar coordinates. The input is a 3-dimensional acceleration vector applied to the quadrotor's center of mass, $\uv = [\ddot{x},\, \ddot{y},\, \ddot{z}]^T$ with a maximum acceleration of $\accel{3}$. The features are squared distances of quadrotor position, $F_1(\x) = \|\myvec{p}_q\|^2$, load's displacement, $F_2(\x) = \|\myvec{\eta}^2\|$, quadrotor's velocity, $F_3(\x) = \|\dot{\myvec{p}}_q\|^2$, and the load's angular velocity, $F_4(\x) = \|\dot{\myvec{\eta}}\|^2$. 
    
	\subsubsection{Learning}  The training domain workspace is volume $\dist{1}$ from the origin, with speed sampled from $[\velocity{-3} \velocity{3}]^{3}$ hypercube. During training the system is deterministic, i.e. there is no stochastic disturbance applied to it. The goal is in the origin. We train with CAVI algorithm. The resulting training weights are $\myvec{\theta} = [-86290\, -350350\, -1430\, -1160]^T.$
	\subsubsection{Planning} The initial condition for aerial cargo delivery are within $\dist{5}$ from the goal. The goal is not in the origin anymore, and the system is influenced by a stochastic disturbance. Input disturbance distributions are evaluated with a mean of 0, 1, and $\accel{2}$, and a standard deviation of 0, 0.5 and $\accel{1}$. 
    \subsubsection{Results}
	\label{sec:res:exp}
    We evaluate LSAPA's 1) suitability for real-time planning of high-dimensional control-affine discrete time systems that require frequent input, 2) ability to execute PBTs in the presence of different stochastic input disturbances, and 3) trajectories for feasibility on physical systems.
    
\begin{figure*}
		\centering
		\begin{tabular}{ccc}
			\subfloat[Quadrotor trajectory]
			{\includegraphics[trim = 0mm 0mm 0mm 0mm, clip,width=0.33\textwidth,height=40mm,keepaspectratio=false]{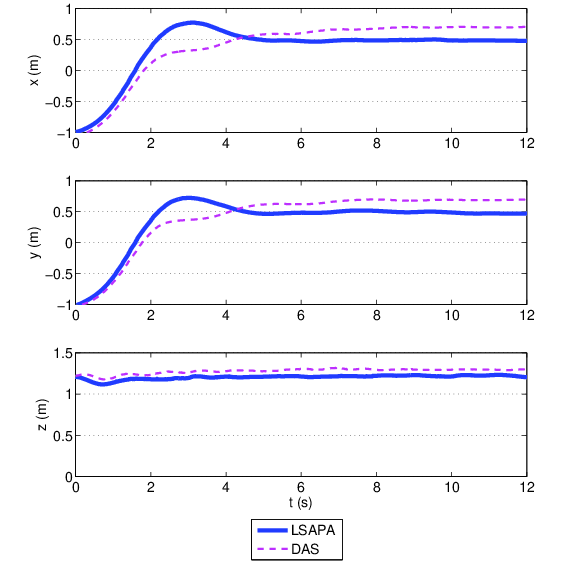}
				\label{fig:quadSF_exp:V}}
			&
			
			\subfloat[Load trajectory]
			{\includegraphics[trim = 0mm 0mm 0mm 0mm, clip,width=0.33\textwidth,height=40mm,keepaspectratio=false]{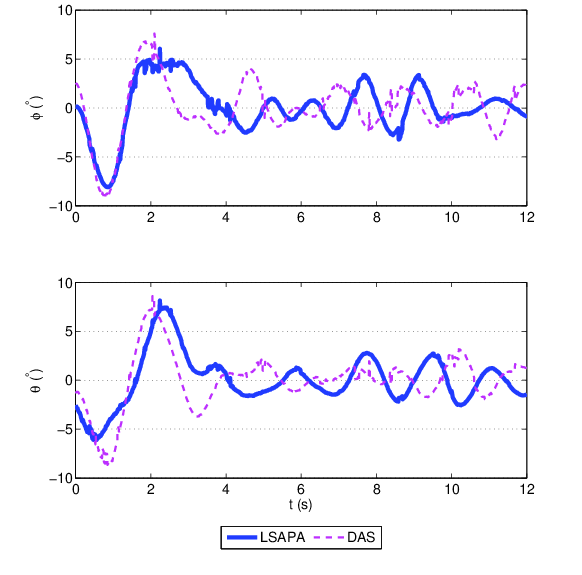} 
				\label{fig:quadSF_exp:S}}
			&
			
			\subfloat[Load trajectory]
			{\includegraphics[trim = 0mm 0mm 0mm 0mm, clip,width=0.29\textwidth,height=40mm,keepaspectratio=false]{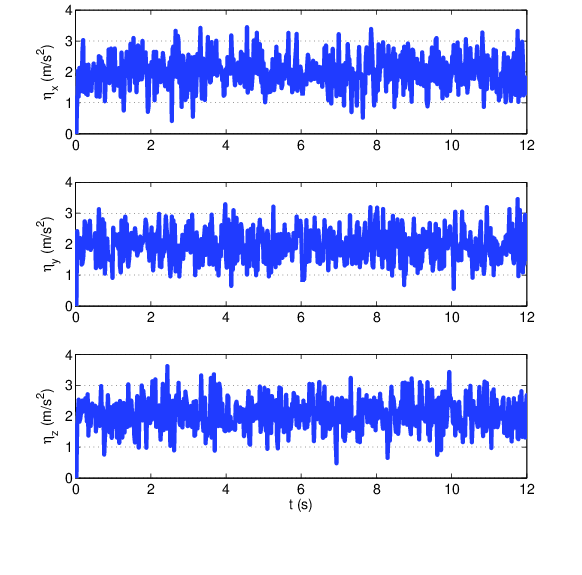}
				
				\label{fig:quadSF_exp:N}}
			
			\\
		\end{tabular}
		\caption{Aerial cargo delivery task: comparison of experimental vehicle (a) and load (b) LSAPA and DAS trajectories with disturbance $\gauss{2}{0.5}$ (c).}
		\label{fig:quadSF_exp}
	\end{figure*}

	To evaluate the feasibility of the LSAPA trajectory  on a physical system and to assess the simulation fidelity, we compare experimentally LSAPA and DAS planned trajectories. We chose DAS for this experiment because it is the fastest input selection method, and it performed better than HOOT in previous experiments \cite{faust-acta-13}. 
	Figure \ref{fig:quadSF_exp} shows the results of the experiment when a disturbance with distribution $\gauss{2}{0.5}$ is imposed on the system (Figure \ref{fig:quadSF_exp:N}). The quadrotor starts at coordinates (-1, -1, 1.2) and the goal is at (0.5, 0.5, 1.2) meters. We notice (Figure \ref{fig:quadSF_exp:V}) LSAPA experiences an overshoot of the goal after 2 seconds, but compensates and returns to the goal position. The DAS trajectory, however, does not compensate and continues with the slow drift past the goal. The load swing is not very different between the two trajectories. 
	
	\begin{figure*}
		\centering
		\begin{tabular}{cc}
			\subfloat[Longitudal]
			{\includegraphics[trim = 0mm 0mm 0mm 0mm, clip,width=0.33\textwidth,height=25mm,keepaspectratio=false]{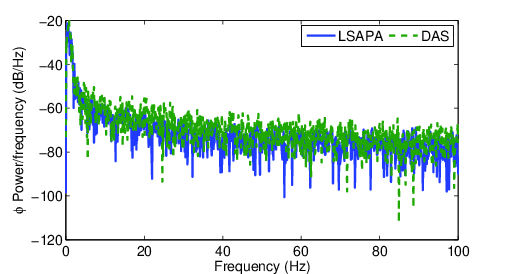}
				\label{fig:quadSF_exp:psd1}}
			&
			
			\subfloat[Lateral]
			{\includegraphics[trim = 0mm 0mm 0mm 0mm, clip,width=0.33\textwidth,height=25mm,keepaspectratio=false]{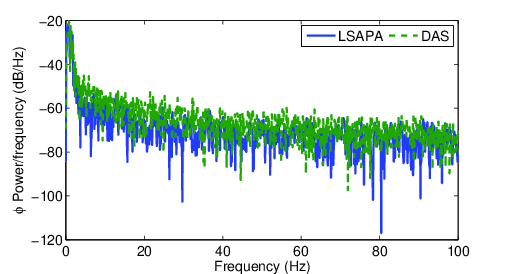} 
				\label{fig:quadSF_exp:psd2}}
			
			\\
		\end{tabular}
		\caption{Power spectral density (PSD) of the suspended load during experimental trajectory created with LSAPA and DAS with disturbance of $\gauss{2}{0.5}$ (c).}
		\label{fig:quadSF_psd}
	\end{figure*}
    We compare LSAPA to DAS, HOOT, and NMPC for the aerial cargo delivery task.  HOOT uses three-level hierarchical search, with each level providing ten times finer discretization. NMPC tracks a trajectory generated assuming no disturbances. It is implemented using the MATLAB routine provided in \cite{nmpc-2011} with a 5 time-step long horizon and cost function $ J(\x, \uv) = \mathrm{E}[\|\myvec{p'} - \myvec{p}_{r}\|^2 + 0.1 \cdot \| \dot{\myvec{p'}} - \dot{\myvec{p}}_{r}\|^2])$, where $\myvec{p'}$  is position of the state that results when input $\uv$ is applied to $\x$. $\myvec{p_r}$ is a position at the reference trajectory. The expectation is calculated as an average of 100 samples. 
	NMPC uses the same disturbance-aware simulator used for LSAPA, DAS, and HOOT. The simulator calculates closed-loop optimization problem and simulates the plan.
    
    Figure \ref{fig:icra15_Table1Replacement} summarizes the planning results. 
	Results in Figure \ref{fig:icra15_Table1ReplacementTime} show that aerial cargo delivery task (left) the time needed to calculate the next input with LSAPA is an order of magnitude smaller than the $20\, ms$ time step (green line), allowing ample time to plan the trajectory in a real-time closed feedback loop. DAS calculates the next input faster than LSAPA. This is expected because the deterministic policy uses 3 samples per input dimension, while the stochastic policy in this case uses 300 samples. NMPC is two orders of magnitude slower for the lower-dimensional aerial cargo delivery task, averaging about $300\, ms$ to select an input, over ten times that the available window for the real-time control. 
	Both LSAPA and DAS are computationally cheap, linear in the input dimensionality, while HOOT, and NMPC scale exponentially. The timing results show that, assuming LSAPA provides good quality trajectories, LSAPA can be a viable method for input selection in real-time on high-dimensional systems that require high-frequency control.
    
	Next, in Figures \ref{fig:icra15_Table1ReplacementDistSW}, we examine if the trajectories complete the task, reaching the goal region of $\distcm{5}$.
	Due to the constant presence of the disturbance, we consider the average position of the quadrotor 
	rather than simply expecting to reach the goal region.
	Note that the accumulated squared error, typically used to measure quality of tracking methods, is not appropriate for LSAPA, HOOT, and DAS because they generate trajectories on the fly and have no reference trajectory. Thus, we measure if the system arrives and stays near the goal. 
	As a control case, we first run simulations for repeatable disturbance ($\gauss{1}{0}$ and $\gauss{2}{0}$). LSAPA, NMPC, and HOOT methods complete the task (Figure \ref{fig:icra15_Table1ReplacementDistSW}).
	
	For a small standard deviation of 0.5 $m / s^2$ LSAPA performs similarly to HOOT, producing trajectories that complete the task, unlike DAS and NMPC. The quality of NMPC solution also degrades with the disturbance (Figure \ref{fig:icra15_Table1ReplacementDistSW}). It is more pronounced than with DAS, because NMPC makes input selection based on solving a fixed horizon optimization problem. The optimization problem accumulates the estimation error, thus invalidating the solution, and smaller horizon lengths are not sufficient to capture good tracking. For the larger standard deviation (1 $m / s^2$), both LSAPA and HOOT create trajectories that still perform the tasks. 
    
\begin{figure*}
		\centering
		\begin{tabular}{ccc}
			\subfloat[Quadrotor trajectory]
			{\includegraphics[trim =  12mm 7mm 17mm 2mm, clip,width=0.33\textwidth,height=45mm,keepaspectratio=false]{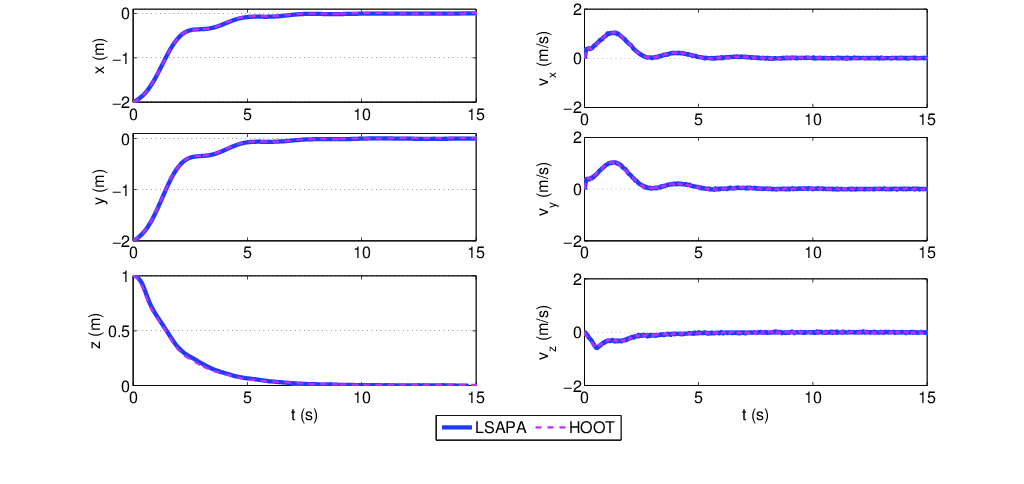}
				\label{fig:quadSF_planning_stochQuadSF_stochAxialSum_N_2_05:V}}
			&
			
			\subfloat[Load trajectory]
			{\includegraphics[trim =  10mm 7mm 17mm 2mm, clip,width=0.33\textwidth,height=45mm,keepaspectratio=false]{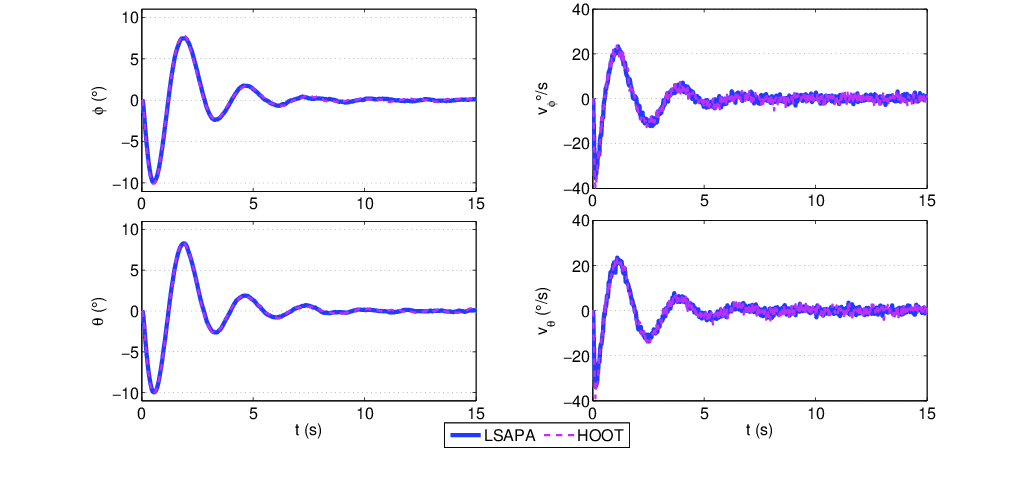} 
				\label{fig:quadSF_planning_stochQuadSF_stochAxialSum_N_2_05:S}}
			&
			
			\subfloat[Injected disturbance]
			{\includegraphics[trim = 12mm 5mm 17mm 2mm, clip,width=0.27\textwidth,height=45mm,keepaspectratio=false]{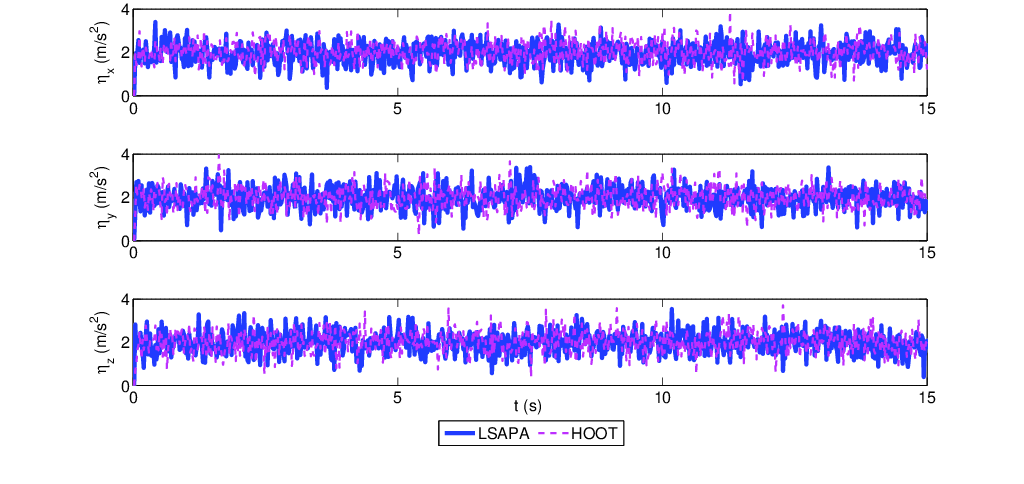}
				\label{fig:quadSF_planning_stochQuadSF_stochAxialSum_N_2_05:N}}
			
			\\
		\end{tabular}
		\caption{Aerial cargo delivery task - comparison of vehicle (a) and load (b) trajectories created with LSAPA and HOOT with disturbance of $\gauss{2}{0.5}$ (c).}
		\label{fig:QuadTrajectoryStoc}
	\end{figure*}

    Figure \ref{fig:QuadTrajectoryStoc} shows trajectories planned with LSAPA and HOOT for aerial cargo delivery task under exertion of disturbance with distribution $\gauss{2}{0.5}$ (Figure \ref{fig:quadSF_planning_stochQuadSF_stochAxialSum_N_2_05:N}). Although both the quadrotor's and the load's speeds are noisy (Figures \ref{fig:quadSF_planning_stochQuadSF_stochAxialSum_N_2_05:V} and \ref{fig:quadSF_planning_stochQuadSF_stochAxialSum_N_2_05:S}), the position changes are smooth, and the quadrotor arrives near the goal position where it remains. 
	
	\subsection{Rendezvous}
	\label{sec:res:rendezvous}
	
	This task is an extension of the aerial cargo delivery, and involves two heterogeneous robots, an aerial vehicle with suspended load, and a ground robot. The two robots work together for the cargo to be delivered on top of the ground robot (Figure \ref{fig:rand}), thus the load must have minimum oscillations when they meet. We use the following definition:
	\begin{mydef} \textit{(Rendezvous task.)}
		A two-robot system, consisting of a ground robot and quadrotor equipped with a suspended load, must navigate from their initial positions towards each other, and come to rest the minimum residual oscillation of the suspended load. Both robots' centers of mass are influenced with a stochastic disturbance with Gaussian distribution.
		The agent has access to the both robots' positions and velocities and the load's position and angular velocity. The goal location is not explicitly known.
	\end{mydef}
    
	\subsubsection{PEARL setup} The state space is a 16-dimensional vector of the joint UAV-load-ground robot position-velocity space, $\x = [\myvec{p}_q,\, \myvec{p}_g,\,\myvec{\eta},\,\dot{\myvec{p}_q},\,\dot{\myvec{p}_g},\dot{\myvec{\eta}}]^T$, and the action set is a 5-dimensional acceleration vector on the UAV (3 dimensions) and the ground robot (2 dimensions), $\uv = [\ddot{\myvec{p}_q},\,\ddot{\myvec{p}_g}]^T$. The maximum acceleration of the UAV is $3\,m / s^2$, while the maximum acceleration of the ground robot is $2\,m / s^2$. Feature vector $\myvec{F}$ contains: $F_1(\x) = \|\myvec{p}_{q_{xy}} - \myvec{p}_{g_{xy}} \|^2 $, the distance between the ground and aerial robot's load $x$ and $y$ coordinates, $F_2(\x) = \| \myvec{p}_{q_{z}} - \myvec{p}_{g_{z}} - 0.6 \|^2 $, the difference in high equal to the suspension cable length, $F_3(\x) = \| \dot{\myvec{p}}_q - \dot{\myvec{p}}_g \|^2 $, their relative speeds, and  $F_4(\x)\| = \myvec{\eta} \|^2$ and $F_5(\x) =\| \dot{\myvec{\eta}} \|^2$ the load's position and velocity relative to the UAV. 
	\subsubsection{Learning} The training samples are sampled from within $\dist{1}$ from the origin. During the training the system is deterministic, i.e. there is no stochastic disturbance applied to it. We learn the task using deterministic CAFVI, which results in the weights $ \myvec{\theta} =   [-92256\, -44767 \, -866\,-336\,-107]^T$.  
	\subsubsection{Planning} After a deterministic learning phase, we generate trajectories for 25 different initial conditions using the learned weights, $\myvec{\theta}$, 
	and varying disturbance distributions.  Rendezvous has the two robots start from within $\dist{10}$ from each other. Input disturbance distributions are evaluated with a mean of 0 and $1\,m / s^2$ and a standard deviation of 0, 0.5 and $1\,m / s^2$, because the maximum acceleration of the ground robot is $2\,m / s^2$ and the system cannot compensate for the larger disturbance. 

	\subsubsection{Results} We compare LSAPA to DAS and HOOT for the rendezvous task. The evaluation set up is the same as in the aerial cargo delivery task.

	\begin{figure*}
		\centering
		\begin{tabular}{ccc}
			\subfloat[Planning time]
			{\includegraphics[trim =  6mm 7mm 16mm 5mm, clip,width=0.25\textwidth,height=45mm,keepaspectratio=false]{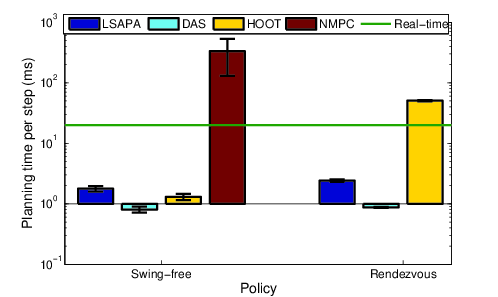}
				\label{fig:icra15_Table1ReplacementTime}}
			&
			
			\subfloat[Aerial cargo delivery]
			{\includegraphics[trim =  7mm 7mm 16mm 7mm, clip,width=0.41\textwidth,height=45mm,keepaspectratio=false]{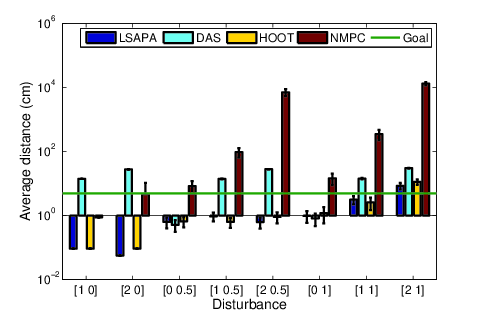} 
				\label{fig:icra15_Table1ReplacementDistSW}}
			&
			
			\subfloat[Rendezvous]
			{\includegraphics[trim = 7mm 7mm 16mm 7mm, clip,width=0.23\textwidth,height=45mm,keepaspectratio=false]{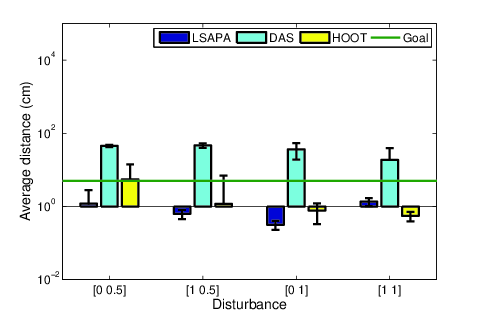}
				
				\label{fig:icra15_Table1ReplacementDistR}}
			
			\\
		\end{tabular}
		\caption{Summary of planning results with LSAPA, DAS, HOOT, NMPC policies for aerial cargo delivery and rendezvous tasks averaged over 25 trials. Time to select a single action (a), and distance from the goal during the last \dur{1} of the flight for aerial cargo delivery (b) and rendezvous (c) tasks. [n, m] signifies disturbance with $\gauss{n}{m}.$ Y-axes are logarithmic. Results below green line are suitable for real-time application (a), and complete the tasks (b) and (c).}
		\label{fig:icra15_Table1Replacement}
	\end{figure*}

Figure \ref{fig:icra15_Table1ReplacementTime} shows that although HOOT performs under $20\, ms$ for the aerial cargo delivery task (left), for the rendezvous task (right), it scales exponentially with the input. As a result, HOOT takes $50\, ms$ to calculate input for the rendezvous task, twice the length of the minimal time step required for real time planning. NMPC for the rendezvous task took 10 times longer than for the aerial cargo delivery task, requiring about 3 hours to calculate a single 15-second trajectory. Thus, we decided against running systematic NMPC tests with rendezvous task because of the impractically long computational time.
	
    Figure \ref{fig:icra15_Table1ReplacementDistR} shows that LSAPA and HOOT trajectories complete the task, unlike DAS and NMPC. While DAS is able to compensate for zero-mean noise for the aerial cargo delivery task (Figure \ref{fig:icra15_Table1ReplacementDistSW}), its performance degrades in the higher-dimensional rendezvous (Figure \ref{fig:icra15_Table1ReplacementDistR}). For rendezvous, LSAPA produces the comparable results in an order of magnitude less time (Figure \ref{fig:icra15_Table1ReplacementTime}).
Overall, LSAPA is the only presented method that performs decision-making in real time, and compensates for the disturbances on the higher-dimensional task. 
    
\begin{figure*}
		\centering
		\begin{tabular}{ccc}
			\subfloat[Quadrotor trajectory]{\includegraphics[trim = 12mm 10mm 17mm 0mm, clip,width=0.33\textwidth,height=55mm,keepaspectratio=false]{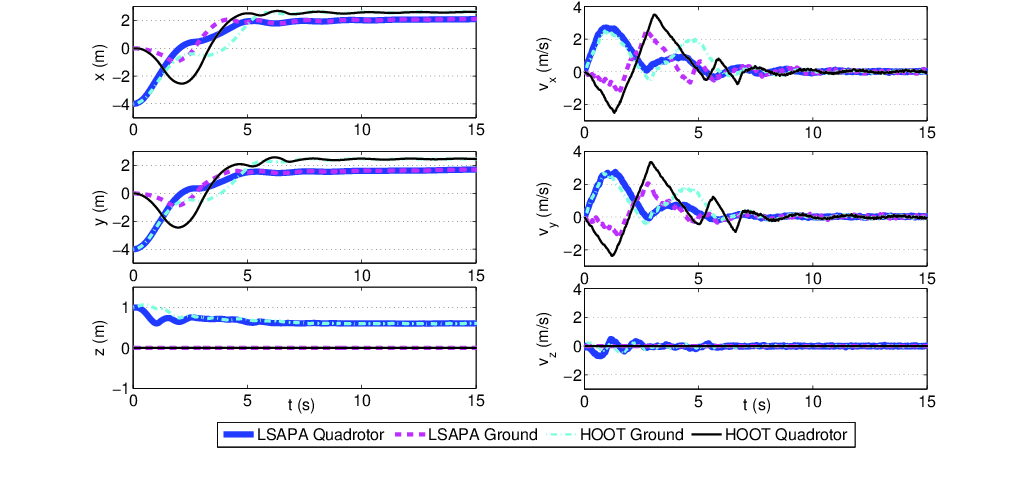}\label{fig:iros14_planning_rand_stochAxialSum_N_1_1:V}}
			&
			
			\subfloat[Load trajectory]{\includegraphics[trim = 12mm 7mm 17mm 0mm, clip,width=0.33\textwidth,height=55mm,keepaspectratio=false]{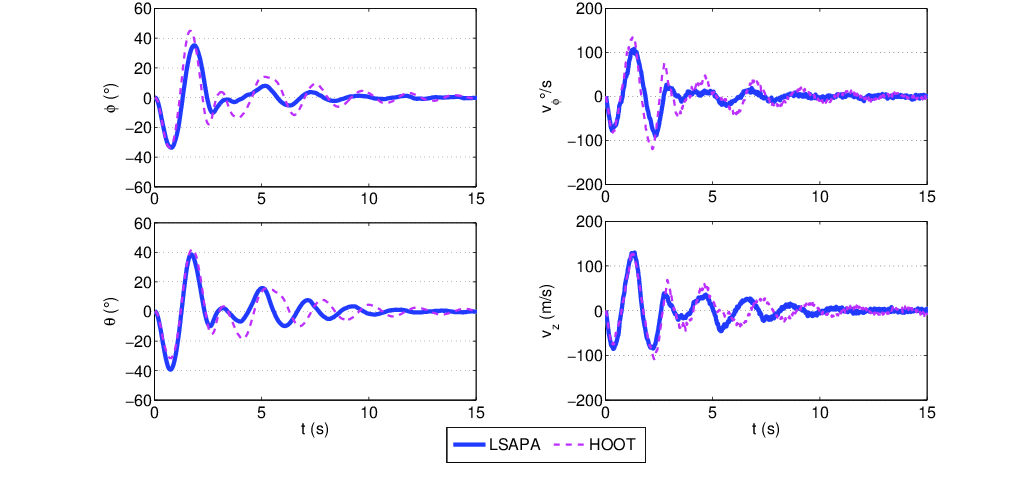}
				\label{fig:iros14_planning_rand_stochAxialSum_N_1_1:S}}&
			
			\subfloat[Injected disturbance]{\includegraphics[trim = 7mm 5mm 12mm 7mm, clip,width=0.27\textwidth,height=55mm,keepaspectratio=false]{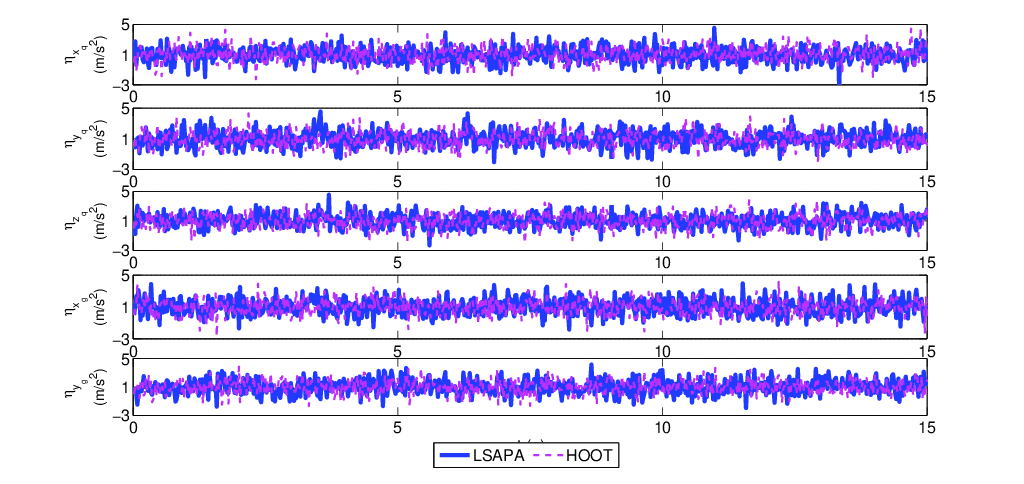}
				\label{fig:iros14_planning_rand_stochAxialSum_N_1_1:N}}\\
			
		\end{tabular}
		\caption{Rendezvous task - comparison of vehicle (a) and load (b) trajectories created with LSAPA and HOOT with disturbance of $\gauss{1}{1}$ (c).}
		\label{fig:RandTrajectoryStoch}
	\end{figure*}

  Figure \ref{fig:RandTrajectoryStoch} show trajectories planned with LSAPA and HOOT for rendezvous task with disturbance with distribution $\gauss{1}{1}$. The two robots meet in 4 seconds, after the initial failed coordinated slow down at 1 second. Note, that the targeted position for the quadrotor's altitude is 0.6 meters in order for the load to be positioned above the ground robot. The results in Figure \ref{fig:icra15_Table1Replacement}
	confirm that both methods produce very similar trajectories, but recall that HOOT does not scale well for larger problems and did not produce rendezvous trajectories in real-time. In contrast, LSAPA produced both trajectories in real-time. 
	
	\subsection{Flying Inverted Pendulum}
	\label{sec:res:flyinv}
	The flying inverted pendulum task consists of a quadrotor-inverted pendulum system in a plane. The goal is to stabilize the pendulum and keep it balanced as the quadrotor hovers \cite{figueroa-wcica-14}. In this paper, we modify the flying inverted pendulum from \cite{figueroa-wcica-14} by additionally applying external stochastic disturbance and extending the planning into 3 dimensional workspace.
    
    \begin{mydef} \textit{(Flying inverted pendulum task)}
		An inverted pendulum is attached to a quadrotor via a massless rigid rod. The quadrotor must balance the inverted pendulum and reduce its own velocity while being influenced by stochastic disturbances.
        The quadrotor has access to the velocity and position of itself as well as the pendulum.
	\end{mydef}
    
    \subsubsection{PEARL setup}
     The state space $\X$ is an ten-dimensional vector of Cartesian position-velocity coordinates, $\x = [\myvec{x_q}\, \dot{\myvec{x_q}} \, \myvec{x_p} \, \dot{\myvec{x_p}}]^T$.
     $\myvec{x_q} \in \R{3}$ is the quadrotor's position in Cartesian coordinates, while $\dot{\myvec{x_q}}$ is its linear velocity. $\myvec{x_p} \in \R{2}$ is the position of the pendulum relative to $\myvec{x_q}$ projected on the plane orthogonal to gravity.
     The action space is a two dimensional acceleration vector $\uv=\ddot{\myvec{x_q}}.$ The maximum acceleration is $\accel{5}$. The reward is one when the target zone is reached, and zero otherwise. The simulator used is a linearized model of a full dynamics of a planar flying inverted pendulum and can be found in \cite{figueroa-wcica-14}. 

    The features for the first task are squares of the pendulum's position and velocity relative to the goal upright position. 
$	\F_{PS}(\x) = [\Vert \myvec{x_p} \Vert^2 \,\, \Vert \mathbf{\dot{\myvec{x_p}}} \Vert^2]^\mathrm{T}.
$
    The second task has an additional feature of a square of the quadrotor's velocity,  
$	\F_{QS}(\x) = [\Vert \myvec{x_q} \Vert^2 \,\, \Vert \mathbf{\dot{\myvec{x_q}}} \Vert^2 \,\, \Vert \myvec{\dot{\myvec{x_p}}} \Vert^2]^\mathrm{T}.$ The subscripts $PS$ and $QS$ denote that just the pole stabilization and quadrotor slowdown tasks are under consideration, respectively.
    
	\subsubsection{Learning}  
    With the exception of the maximum acceleration, the training set up is the same as in \cite{figueroa-wcica-14}, and we use DAS policy in training. The resulting parametrization vectors are $\teta_{PS}$ = $[-86.6809   -0.3345]^T$ and $\teta_{QS}$ = $10^6[-1.6692, -0.0069, 0.0007]^T $. 
	
	\subsubsection{Planning} We use a disturbance probability density function $\gauss{1}{1}$ and a pole initial displacement of  23$^\circ$. While the deterministic sum solves this problem and balances the inverted pendulum in the absence of disturbances and small zero-mean disturbances ($\gauss{0}{0.5}$), it fails to balance the inverted pendulum for non-zero mean disturbances. In contrast, LSAPA policy solves the task (Figure \ref{fig:InvPendTrajectoryStoch}). 
    
    \subsubsection{Results}
\begin{figure*}
		\centering
		\begin{tabular}{cc}
			\subfloat[Quadrotor trajectory]{\includegraphics[trim = 7mm 0mm 10mm 0mm, clip,width=0.47\textwidth,height=35mm,keepaspectratio=false]{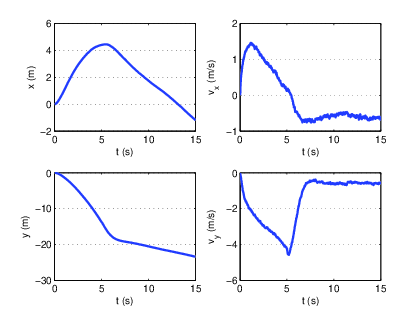}\label{fig:iros14_invPend_stochAxialSum_1_1_V}}
			&
			\subfloat[Inverted trajectory]{\includegraphics[trim = 5mm 0mm 10mm 0mm, clip,width=0.47\textwidth,height=35mm,keepaspectratio=false]{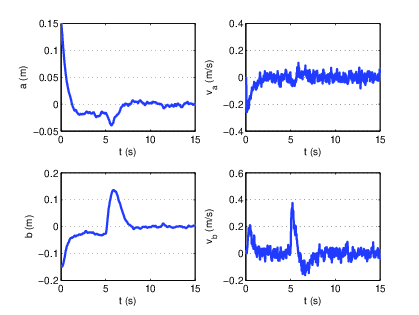}
				\label{fig:iros14_invPend_stochAxialSum_1_1_L}}\\
			
		\end{tabular}
		\caption{Flying inverted pendulum trajectory created with LSAPA with disturbance of $\gauss{1}{1}$.}
		\label{fig:InvPendTrajectoryStoch}
	\end{figure*}

    Figure \ref{fig:iros14_invPend_stochAxialSum_1_1_V} shows the quadrotor's trajectory, and Figure \ref{fig:iros14_invPend_stochAxialSum_1_1_L} displays pendulum position in Cartesian coordinates relative to the target position above the quadrotor. The first subtask brings the pole upright (0 to 5 seconds). Then the second subtask slows down the quadrotor (after 5 seconds). The pole is slightly disturbed during the initial moments of the second subtask but returns to an upright position.
	
\begin{figure*}
		\centering
		\begin{tabular}{ccc}
			\subfloat[Accumulated reward]{\includegraphics[width=0.3\textwidth,height=25mm]{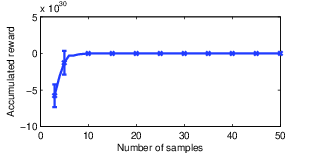}\label{fig:iros14_invPend_reward}}
			&
			\subfloat[Displacement]{\includegraphics[width=0.3\textwidth,height=25mm,keepaspectratio=false]{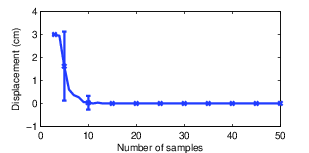}
				\label{fig:iros14_invPend_displacement}}&
			\subfloat[Speed]{\includegraphics[width=0.3\textwidth,height=25mm,keepaspectratio=false]{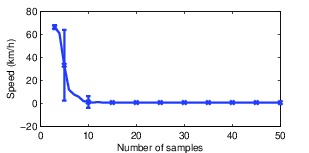}
				\label{fig:iros14_invPend_speed}}
		\end{tabular}
		\caption{Trajectory characteristic per sample size in flying pendulum with disturbance $\gauss{1}{1}$, using LSAPA; mean and standard deviation over 100 trials.}
		\label{fig:InvPendSamples}
	\end{figure*}

	Figure \ref{fig:InvPendSamples} depicts the results of the trajectory characteristics for increasing number of samples in LSAPA. 
	The smallest number of samples is three. The accumulated reward (Figure \ref{fig:iros14_invPend_reward}) increases exponentially below 10 samples.  The gain decreases between 10 and 20 samples. Thus, the peak performance is reached after 20 samples. Sampling beyond that point brings no gain. We see the same trend with the pole displacement (Figure \ref{fig:iros14_invPend_displacement}) and speed magnitude (Figure \ref{fig:iros14_invPend_speed}).
	
	\section{Discussion}
	\label{sec:discussion}
	
	In this section we analyze and discuss \frm's policy progression to the attractor, computational complexity, preference properties, and training domain selection. First, we find sufficient conditions for the policy to progress to the attractor, and analyze the regions where the agent can get stuck when the conditions are violated. Second,  we prove that the LSAPA computational complexity is linear with the dimensionality of the input space. Third, we consider the preferences' properties with respect to the workspace volume and dimensionality. Last, we construct the training domain which captures both the interesting parts of the problem, and is small enough for the efficient learning. 
	
	\subsection{Task completion} 
	In this section we discuss the policy's progression to the attractor. First, we look at convergence conditions for attractor-only tasks, and then discuss the existence of local maxima when the task contains repellers. 
	\subsubsection{Attractor-only tasks}
	
	A greedy policy \eqref{eq:greedy} with respect to a state-value function V \eqref{eq:value} progresses to the attractor for a control affine system that satisfies the following conditions:
	\begin{enumerate}
		\item The system is controllable and the attractor is reachable. In particular, we use,
		\begin{equation}
		\label{eq:controllable}
		\exists i, 1 \leq i \leq \nr,\text{ such that } \bx\Lvp\g_i(\x) \ne 0, 
		\end{equation}
		and that $\A$ is regular outside of the origin,
		\begin{equation}
		\label{eq:regular}
		\A^T\Lvp \A > 0\text{, } \x \in \X \setminus \{\myvec{0} \}
		\end{equation}
		\item Action is defined on a closed interval around the attractor,
		\begin{equation}
		\label{eq:action}
		\myvec{0} \in \U
		\end{equation}
		\item The drift is bounded,
		\begin{equation}
		\label{eq:drift}
		\bx^T\Lvp \bx \leq \x^T\Lvp \x, \text{ when } \Lvp > 0
		\end{equation}
		\begin{equation}
		\label{eq:V2}
		V(\x) = \x \Gamma \x^T, \Gamma < 0
		\end{equation}
		
	\end{enumerate}
	
	The claim is the direct result from Artstein's theorem \cite{artstein}, and our previous work \cite{faust-ai-13-journal,faust-acta-13}.   Artstein's theorem \cite{artstein} states that a greedy policy over a control Lyapunov function takes the system to the equilibrium. And in  \cite{faust-ai-13-journal,faust-acta-13} we showed that under the conditions above, a control Lyapunov function can be constructed from the state-value function V.
	
	When the learning state-value function V is in the form  \eqref{eq:V2}, the greedy policy \eqref{eq:greedy} will progress to the equilibrium, located in the goal.  The goal preferences result in a state-value function of the form \eqref{eq:V2} when the objectives span the entire state space and each element of the learned vector $\theta$ is negative. To show why, we write the goal preferences using the projection notation.
	
	Let $P_k $ be a $2d_r\,\text{x}\,2d_r$ diagonal projection matrix corresponding to the attractor $\obj{k} \in P_a$, $P_k = \text{diag}(p_j | j = 1,..,2d_r, p_k = id(\obj{k,j} \ne 0)).$ The $j^{th}$ element on $P_k'$s diagonal is equal to 1, only if $\obj{k}'$s $j^{th}$ coordinate is non-zero. All other elements are 0. Clearly, $P_k \ge 0$ for $k = 1,..,\no.$ Then
	\begin{align*}
		F_k(\x) &= (P_k\x - \obj{k})^T(P_k\x - \obj{k}) \\
		&= P_k(\x-\obj{k})^TP_k(\x-\obj{k}) \\
		&= (\x-\obj{k})^TP^T_kP_k(\x-\obj{k})
		&= (\x-\obj{k})^TP_k(\x-\obj{k}),
	\end{align*}
	because $P_k=P_k^T P_k$ since $P_k$ is an orthogonal projection matrix. Let $P = \sum_{k=1}^{\no} P_k.$ $P$ is also a diagonal, orthogonal, and positive semi-definite. Further, $P > 0,$ when the objectives span the whole state space, i.e. when $rank (P) = 2d_r.$ Similarly, for $P_\theta = \text{diag}(\theta),$ matrix $P_\theta P < 0$ iff $P >0$ and each coordinate of $\theta$ is negative. Since \eqref{eq:value} can be written as
	\begin{align*}
		V(\x)&= \theta^T (\x-\obj{})^TP(\x-\obj{}) \\
		&= (\x-\obj{})^T\text{diag}(\theta)P(\x-\obj{}) \\
		& = (\x-\myvec{\obj{}})^TP_{\theta}(\x - \obj{}) = \x_{\obj{}}^TP_{\theta}\x_{\obj{}}, 
	\end{align*}
	after translation of the coordinate system by $-\obj{},$ where $\obj{} = \sum_{k=1}^{\no} \obj{k}.$ 
	
	To conclude, the policies learned with \frm\ for tasks with only attractor that jointly span the entire state-space are guaranteed to progress the agent to the goal $g = \cup_k \obj{k}$ when the learning algorithm results in parametrization $\theta$ with all negative coordinates.
	
	\subsubsection{State value function local minima analysis}
	For tasks with mixed preferences, such as dynamic obstacle avoidance, the agents follow
	preferences, but there are no formal completion guarantees. In fact, the value
	function \eqref{eq:value} has potentially two maxima, one on each side of the
	obstacle. 
	
	For the purpose of this analysis, we assume that the problem is well formulated, and
	contains one goal, 
	\ie the intersection of subspaces defined by distance reducing objectives is 
	non-empty, and forms a connected set. Note that in order
	for both attractor and repeller as expected 
	the resulting weights must be negative $\theta_i < 0$. 
	Since a straight line is the shortest path between an agent and its attractor, we
	analyze the value function restricted to that line with varying obstacle
	distances. To simplify the analysis, we transform the value function \eqref{eq:value}. 
	Without loss of generality, we rotate and scale the
	coordinate system, such that the goal is in the origin, and the agent is on
	the $x-$axis. The two nearest obstacles lay on (1, $d$), and (1, $-d$). In addition, we multiply
	the entire function \eqref{eq:value} by minus one, to give a rise to function $V_x(x)$. Now, we are interested
	in finding necessary conditions $V_x(x)$ minima, which correspond to the $V(x,y)$ maxima. 
	
	First to construct $V_x(x)$, let $c = \frac{\theta_2}{\theta_1} > 0$ be the
	ratio between learned weights for the repeller and the attractor feature. The value
	function after the affine transformation is $$V_x(x) = -1 * V(x,0)= x^2 +
	\frac{c}{(x-1)^2 + d^2}.$$ We examine necessary conditions for $V_x(x)$'s minima based on the obstacle distance, $d,$ and the coefficient $c.$ Point $x_0$ is local minima if
	\begin{equation}
	\label{eq:dv}
	\frac{\text{d}{V_x}}{\text{d}x}(x_0) = 2x_0 - \frac{2c(x_0-1)}{{((x_0-1)^2 + d^2)^2}} = 0 ,
	\end{equation}
	and the second derivative is positive,  
	\begin{equation}
	\frac{\text{d}^2{V_x}}{\text{d}{x}^2}(x_0) = 2 + 2c\frac{(x_0-1)^2 - d^2}{{((x_0-1)^2 + d^2)^3}} > 0.
	\label{eq:ddv}
	\end{equation}
	For \ref{eq:dv} to hold, the following must be the case,
	\begin{align}
		\label{eq:dv1}
		x_0 < -1, \text{ or } x_0 > 0,
	\end{align}
	and  given \eqref{eq:dv1}, \eqref{eq:ddv} holds when,
	\begin{equation}
	\label{eq:ddv1}
	\|x_0 - 1 \| > \|d\|. 
	\end{equation}
	We conclude that the value function \eqref{eq:value} for problems with obstacles has two local maxima, one to the left of the goal, and the other one to the right of the obstacles. If the agent is located between the goal and the obstacles, it will settle at the equilibrium point to the left of the obstacle. Otherwise, the agent will settle to the right of the obstacle, unless the obstacles move. Further, the equilibrium point tends to the goal, as the distance between the obstacles increase. 

\begin{figure}[h]
		\includegraphics[trim=0mmm 0mm 0mm 0mm,
		clip,width=\linewidth,height=3.8cm,keepaspectratio=false]{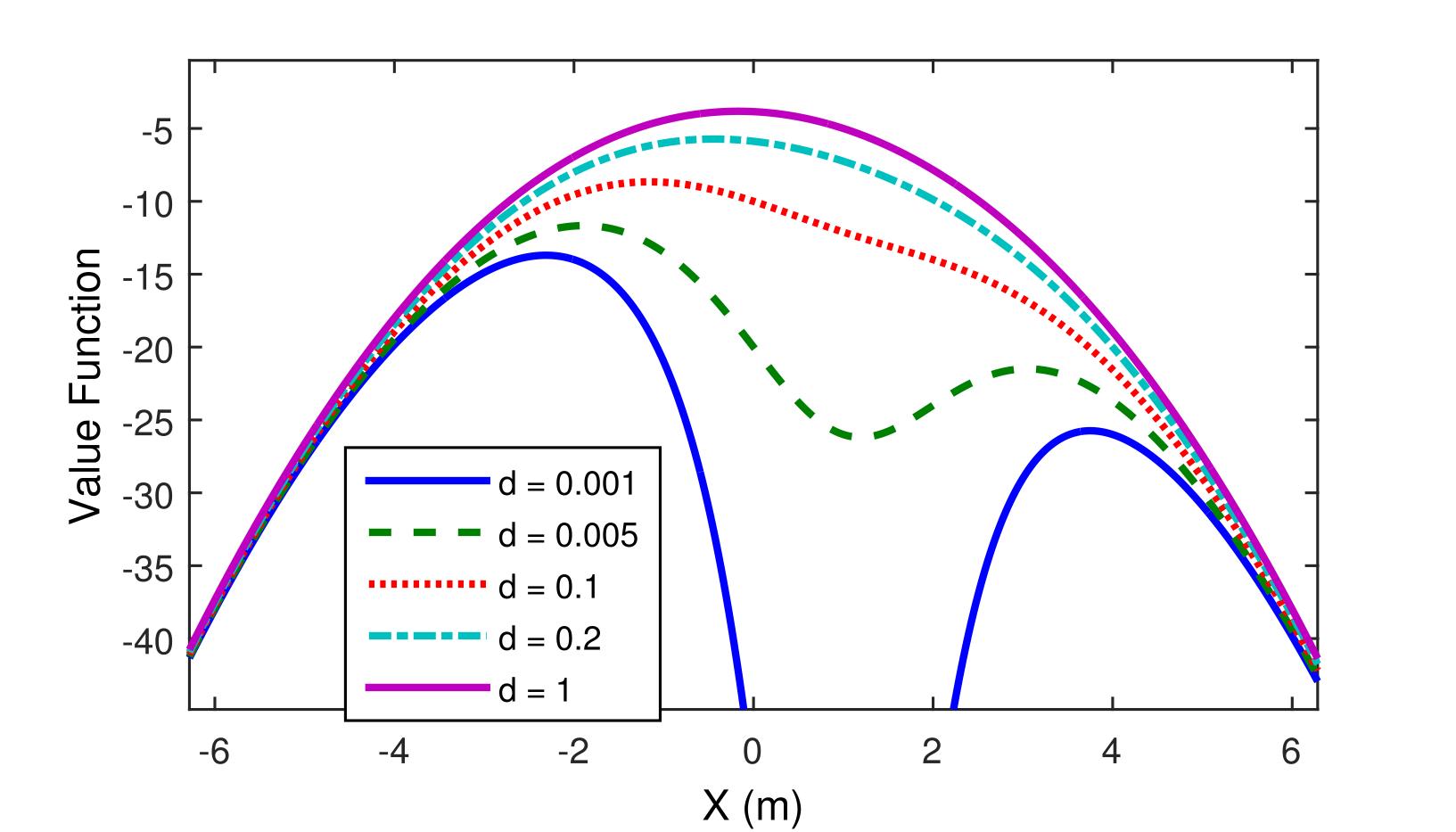} 
		\centering
		\caption{\small Value function inflection points for $c=100$.}
		\label{fig:analysis}
	\end{figure}

	We perform an empirical study, depicted in Figure \ref{fig:analysis}, to verify the analysis.  The value function has either a single maxima near an attractor (pink, and blue lines in Figure \ref{fig:analysis}), has two maxima (blue and green line), or has an inflection point near the obstacles (red line). 
	Inspection of the partial derivative $\pder{V}{y}$ at the minima points in
	Figure \ref{fig:analysis} reveals that these points are saddle points. 
	
	In summary, when the obstacles are far enough apart there is only a global maximum. As
	the obstacles come closer together, a new region of attraction forms on the other
	side of the obstacle. If the agent gets into the local maximum region of attraction,
	gradient-descend methods trap it. Sampling-based greedy methods such as
	HOOT, however, might get the agent out of the local maxima if it is close enough to the boundary.  

	\subsection{Computational cost}
	
	The feature vector computation time given in \eqref{eq:f}  is linear in state dimensionality and number of preferences. This is because 
	\[\oo{\F(\x,\nx)} = \sum_{i=1}^{\no}\oo {\F_i(\x,\nx)} = \oo{\no  \F_i(\x,\nx)} = \oo{\no \nx},\]because projection is linear operation.

	\begin{proposition}
		\label{th:lsaspa}
		The computational cost to calculating LSAPA with \eqref{eq:lsaspa} is $\oo{\no \cdot \nr \cdot \nx^2 \cdot d_n}$,  assuming that $\no \leq \nr \leq d_x \ll d_n$.
	\end{proposition}
	The proof is in Appendix \ref{sec:app:2}.
	Thus, LSAPA's running time depends on the state dimensions instead of their physical size. On the other hand, its running time depends on the number of preference, $\no$, and samples, $d_n$.

	\subsection{Preference properties} 
	Now we discuss the feature properties, and how they handle changes in the workspace volume and dimensionality.
	
	The preferences project the state space $\X$ onto a $\nx$-dimensional manifold in
	the lower-dimensional preference space. 
	Each preference $\Fi_i$ is a Lipschitz continuous function defined for the entire state space, $\Fi_i:\R{m} \rightarrow \R{},\;i=1,..,\nx$. They are monotone, with a single global extreme (maximum for attractors, and minimum for repellers).  
	$\Fi_i:$ projects $(\R{m}-1)-$spheres, centered at the corresponding points of interest, $\obj{i},$ to scalars. The projection defines an equivalency kernel of $\F_i$, $=_{F_i}$, between states $\x_1, \x_2 \in \X $ such that 
	$$ \x_1 =_{F_i} \x_2 \,\,\text{iff}_{def} \,\, \F_i(\x) = \F_i(\myvec{y}).$$
	
	The state-value function \eqref{eq:value} is approximated  with a weighted sum of $2\nx$-dimensional smooth,  equivalency kernel in the preference space. The resulting function is Lipschitz continuous, although, as we've seen earlier it has number of local minima and maxima. The number of features does not change with the physical size and dimensionality of the workspace. The computation time grows linearly with the dimensionality of the workspace, as we showed in the previous Section.
 
	State-space tiling methods \cite{wu-largescale-10} also form equivalency kernels. However, the projection functions are defined only on the tile, and the resulting state-value function has discontinuities. Further, the number of equivalence classes grows with state space volume, and the growth is exponential with the number of dimensions. The preferences, presented here, are a form of radial-bias functions \cite{BusBab:10-002}, in that they are smooth, monotone, defined on the whole state space domain, and the state-value function is a weighted sum. The differentiating factor is that the classical radial-bias methods anchor the bases equidistantly, while we anchor them in the task's goals and obstacles. Because they are anchored equidistantly, their number grows the same as with the tiling approaches. Neural net approximators using convolution layers \cite{tamar2016value} \cite{atari-paper}  have a fixed number of features that do not change with the state space size. They discover the transformation from the state to task feature space through the training, without taking advantage of knowledge of the task objectives, requires much larger sample size and longer training.
	
	The planner solves problems with larger state and action domains than it was trained on. The observed state $\x \in \X$, and recommended action $\uv \in \U$ are from the full problem MDP  $\mathcal{M}$ defined in \eqref{eq:mdp}. This is because the features are defined and valid in $\mathcal{M}$, because they capture the important elements of the task rather than the physical space.  The features enable both efficiency, by learning on small problems, and adaptation, by allowing the
	policy transfer to larger problems. 
	It is the use of the features that separates \frm\ from standard RL, and creates an easy to use  task learning and planning method.
	
	\subsection{Learning domain selection}
	This section discusses two approaches for selecting the domain for efficient training. 
	First, for the training to produce good results, the training samples need to capture the problem space well. We discussed earlier that the larger state space is projected onto the lower dimensional task space. And, the analysis of the local mininma showed that all the state-value functions' local maxima are located in the vicinity of the attractors and repellers. The bounded segment that contains all attractors and repellers will contain all interesting training samples. We can restrict the learning domain to that area.
	
	Second, to reduce the learning domain without affecting the training quality, we can reduce the workspace dimensionality (states and actions). \frm\ learns in the quotient set of the equivalency kernel, which is smaller. By reducing the size of each partition, we increase the coverage of the quotient set during the training under a fixed sampling budget. 
	
	For example, in the motion planning cases that we consider for the quadrotors, the 2-dimensional workspace captures the complexity of the problem while the quadrotor operates in the 3-dimensional workspace. The 3-dimensional workspace increases the volume of the learning domain without exposing additional problem complexity. Thus, we can further simplify the learning problem and look only at the 2-dimensional projection, reducing the dimensionality of both state space $\X_l \subset \R{4{\nr}} \subseteq \X$ and $\U_l \subset \R{2{\nr}} \subseteq \U.$ 
	
	In another example, consider the rendezvous task (Figure \ref{fig:rand}), which is a task in a 3-dimensional workspace 10 by 10 meters, leading to 10-dimensional state space and 5-dimensional action space. Assuming we limit the velocity to 3 meters per second per dimension, the volume of the planning space is $3^5 10^5 = 24\,300\,000.$ We can train the rendezvous task in 1 meter by 1 meter 2-dimensional workspace, and limiting robot acceleration to 1 meter per second squared, resulting in an 8-dimensional state space with the volume of 1. The training action domain $A_l$ is a 4-dimensional segment. It is clear that given a fixed sampling budget to train AVI, a better coverage can be achieved in the reduced learning domain, than on the original full planning domain.
	
	\section{Conclusion}
	\label{sec:conclution}
	This paper presents \frm , an efficient, motion-planning training and planning tool for multi-agent, high-dimensional, preference balancing tasks, appropriate for obstacle avoidance and point-to-point navigation with control-affine systems. By expressing the tasks as a combination of user attractor and repeller intents, the method learns to balance the features, and solves tasks that could be challenging to solve otherwise. We presented feature construction from user intents, formulation of the learning MDP to speed up the training, and adaption of the method to compensate for stochastic disturbances during planning. The method is demonstrated on multi-agent pursuit and dynamic obstacle avoidance tasks, where we train in small static environments. We also demonstrated the stochastic disturbance extension on the aerial cargo delivery problem, rendezvous, and flying inverted pendulum problems, including experimentally on a physical quadrotor. Lastly, we derived the conditions for guarantees for progression to the goal, discussed computational complexity of the policies, and analyzed the preference properties and training domain reduction. Overall, \frm\ is easy to use, fast to train, and although it offers limited convergence guarantees, it is applicable for a variety of motion planning problems.
	
	\appendices
	\section{Proof of the Proposition \ref{th:prop}}
	\label{sec:app:1}
	\begin{proof}
		The proof is similar to the proof of Proposition 3.1 in \cite{faust-acta-13} and relies on the fact that although the stochastic disturbance changes the system dynamics at every time step, it does not interact with the control input.
		
		After rearranging terms we can write the dynamics \eqref{eq:dynSystem} as 
		\[
		\D: \;\;\; \x_{k+1} = \bnx{k} + \myvec{g}(\x_k)\uv_k,
		\]
		where $\bnx{k} = \myvec{f}({\x_k}) + \myvec{g}(\x_k)\noise{k}$ does not depend on input $\uv_k$.
		Let us denote $\Lambda = \C \myvec{\Theta} \C^T$. For an arbitrary state $\x$ and any value of the disturbance $\noise{}$,
		\begin{align}
			Q(\x, \uv, \noise{}) &= V(\myvec{D}(\x, \uv, \noise{})) \\
			& = V(\bnx{} + \A\uv ) \\
			&=(\bnx{} + \A\uv ))^T\Lambda (\bnx{} + \A\uv ).\label{eq:q1} 
		\end{align}
		Thus, $Q$ is a quadratic function of action $\uv$ at any fixed state outside the origin, $\x \in \X \setminus \{\myvec{0}\},$ and fixed disturbance $\noise{}$.
		
		To show that $Q$ has a maximum, we inspect $Q$'s Hessian for fixed state $\x$ and disturbance $\noise{}$,
		\begin{align*}
			HQ(\x, \uv, \noise{}) &= \begin{bmatrix}
				\ppder{Q(\x, \uv, \noise{})}{u_1}{u_1} & ... & \ppder{Q(\x, \uv, \noise{})}{u_1}{u_{\na}}\\ 
				& ... & \\ 
				\ppder{Q(\x, \uv, \noise{})}{u_{\nr}}{u_1} & ... & \ppder{Q(\x, \uv, \noise{})}{u_{\nr}}{u_{\na}} 
			\end{bmatrix}\\
			&= 2\myvec{g}(\x)^T\Lambda \myvec{g}(\x),	
		\end{align*}
		which cancels the stochastic term, because the stochastic term does not affect square of the input $\uv$ as seen in \eqref{eq:q1}. Because $\A$ is regular for all states $\x \in \X \setminus \{\myvec{0}\}$ and $\myvec{\Theta} < 0$, the Hessian is negative definite, so $Q$ is concave with a maximum for an arbitrary state outside of the origin.
	\end{proof}

	\section{Proof of Proposition \ref{th:lsaspa}}
	\label{sec:app:2}
	\begin{proof}
		The complexity of calculating $\x'_{j,i}$ for one-dimensional input is $\oo{\nx}$. Since $\F$ can be calculated in $\oo{\nx}$, the complexity of \eqref{eq:qi} is $\oo{\no \cdot d_n \cdot \nx^2}$. Formulating matrix $C_i$ in \eqref{eq:ci} is $\oo{d_n}$. The solution to the regression problem \eqref{eq:reg} and \eqref{eq:regRes} is asymptotically $\oo(d_n),$ since we use polynomial of the second degree for the regression. Thus the asymptotic complexity of calculating \eqref{eq:hatu} is $\oo{\no\cdot  d_n \cdot \nx^2} +\oo{d_n}+ \oo{d_n} = \oo{\no \cdot d_n\cdot  \nx^2}$. Finally, the complexity of the final input selection \eqref{eq:max} is $\oo{\no \cdot \nr \cdot \nx^2\cdot  d_n}$ after repeating the process for all axes. 
	\end{proof}
	
	\section*{Acknowledgment}
	
	The authors thank Peter Ruymgaart for helpful discussions on modeling the external disturbances, Particio Cruz for assisting with experiments, Angela Schoellig and Ivana Palunko for very useful feedback and discussions of Model Predictive Control, and John Baxter and Marco Morales for the feedback on the early versions of the manuscript.  We thank Jur Ven Den Berg for feedbacks on modifying ORCA. PEARL was developed by Faust while at University of New Mexico, and partially supported with New Mexico Space Grant. 
	Tapia and Chiang are supported by the National Science Foundation under Grant Numbers IIS-1528047 (NRI) and IIS-1553266 (CAREER).  
	Any opinions, findings, and conclusions or recommendations expressed in this material are those of the authors and do not necessarily reflect the views of the National Science Foundation. 
	\bibliographystyle{IEEEtran}
	\bibliography{literature}
	
\end{document}